%% file: main.tex
\documentclass{article}

\usepackage[final]{corl_2026} 

\input{preamble}
\usepackage{wrapfig}
\PassOptionsToPackage{numbers,sort&compress}{natbib}
\usepackage[normalem]{ulem}

\title{\modelname: A Single Backbone for Video \\ Generation and Robot Control}

  \author{%
    Jeremy A.~Collins\textsuperscript{1,3,\dag}
    \And
    Seungjae Lee\textsuperscript{2,3,\dag}
    \AND
    Rachanon Wachakorn\textsuperscript{1}
    \And
    Krishnan Srinivasan\textsuperscript{3}
    \And
    Animesh Garg\textsuperscript{1}
    \AND
    Vitor Campagnolo Guizilini\textsuperscript{3,\ddag}
    \And
    Paarth Shah\textsuperscript{3,\ddag}
  }

\newcommand{\modelname}{\textsc{Tapestry}\xspace}

\begin{document}

\maketitle

  \renewcommand{\thefootnote}{}
  \footnotetext{\textsuperscript{1}Georgia Institute of Technology. \quad \textsuperscript{2}University of Maryland. \quad \textsuperscript{3}Toyota Research Institute. \\
  \textsuperscript{\dag}Work performed partially during internship at Toyota Research Institute. \quad \textsuperscript{\ddag}Equal advising.}
  \renewcommand{\thefootnote}{\arabic{footnote}}

\begin{abstract}
Pretrained video generation models encode rich priors about physical interactions, but leveraging them for robot control typically requires either resource-intensive video generation at test time or action heads trained from scratch that do not fully utilize the pretrained backbone. We introduce Tokenizing Actions as Video Patches in a Shared Transformer Policy (TAPESTRY), a world action model that repurposes a pretrained video diffusion transformer into a robotic policy without requiring video generation at inference. Our key idea is to treat robot actions as single-patch tokens in the video model's latent space. A lightweight linear projection maps each action timestep to the patch dimensions of the pretrained VAE, and the resulting tokens are processed by the same patchifying layers and transformer backbone used for video. The shared backbone is optimized through three diffusion objectives sampled per minibatch element:
\textit{Video diffusion} uses VAE latents from robot and human videos to establish dynamics priors. \textit{Joint video+action diffusion} trains on robot demonstrations using a coupled diffusion timestep and an asymmetric attention mask. \textit{Action-only diffusion} denoises actions conditioned on clean image latents. At test time, the model generates a compact action token sequence in a single DDIM step from pure noise, producing smooth action chunks at $84$\,ms per action chunk on consumer hardware and avoiding the cost of full video generation. Across simulation benchmarks and real-robot experiments, TAPESTRY achieves strong performance with substantially lower inference cost than comparable joint video--action policies. More information can be found at \href{https://tapestry-wam.github.io}{tapestry-wam.github.io}.

\end{abstract}

\section{Introduction}

Video generation models have emerged as an effective method of distilling internet-scale knowledge about physical interactions into robotic policies.
Trained on up to millions of hours of human and real-world video, these models learn rich priors over how the visual world evolves in response to actions, priors difficult to acquire from the limited robot demonstration datasets available today.
The central question for robot learning is: \textit{how can we best transfer these priors to learn policies that generalize across tasks, environments, and embodiments?}

Existing approaches span a spectrum of coupling between the video model and the robotic policy, each with characteristic tradeoffs.
At one end, methods such as UniPi~\citep{du2023learning} and DreamGen~\citep{jang2025dreamgen} use a video generation model to synthesize visual rollouts and then extract actions via a separately trained inverse dynamics model.
This pipeline leaves the video model architecturally unmodified but compounds errors across two generative stages and requires expensive frame generation at test time, limiting real-time deployment.
A lighter-weight alternative, exemplified by Video Prediction Policy~\citep{hu2024video} and mimic-video~\citep{zhong2025mimicvideo}, extracts intermediate latent representations from a pretrained video backbone and conditions a downstream action head on these features.
This avoids full video generation but limits knowledge transfer, since the action head is trained from scratch and cannot directly access the backbone's pretrained weights.
The most tightly coupled approaches share a single backbone across video and action diffusion. These include prior work UWM~\citep{zhu2025unified}, UVA~\citep{li2025unified}, and concurrent works Cosmos Policy~\citep{kim2026cosmospolicyfinetuningvideo}, DreamZero~\citep{ye2026dreamzero}, Fast-WAM~\citep{yuan2026fastwamworldactionmodels}, and GigaWorld-Policy~\citep{ni2026gigaworld}.
These methods achieve the deepest integration, but each either trains from scratch without a pretrained video prior, pays a heavy inference-time token cost, or predicts actions with a separate module trained from scratch.

Our insight stems from a simple observation: the dynamics priors useful for action prediction reside in the \emph{weights} of the video backbone, not in the generated pixels.
Although generating full video sequences is one way to access these priors, we identify that it is unnecessary if the end goal is fast, closed-loop action prediction.
Specifically, by representing action tokens as patchified video tokens aligned with video pretraining, robot actions inherit the backbone's dynamics priors without the overhead of pixel-space generation, reducing online inference cost.

We apply this idea in \modelname, a method that re-purposes a pretrained video diffusion transformer into an action-generating policy.
The key mechanism is treating each action timestep as a single patch in the video VAE latent space: a lightweight linear projection maps the raw action vector to the VAE's patch dimensions, an affine-free layer normalization matches the per-token mean and variance of the VAE latents, and the resulting pseudo-patch is fed through the \emph{same} patchifying convolution that the backbone uses for video tokenization.
The backbone is then jointly fine-tuned on three diffusion objectives sampled per minibatch element: standard video diffusion on VAE latents from a large mixture of robot and human videos, joint video$+$action diffusion on paired robot demonstrations, and action-only diffusion that denoises actions conditioned on clean image latents.
At test time, only the action pathway is used: the model denoises a compact sequence of action tokens conditioned on the current observation and a language instruction, in a single DDIM step from pure noise.
This produces smooth action chunks at $84$\,ms per chunk on a single RTX~5090, ${\sim}44\times$ faster than an optimized DreamZero~\citep{ye2026dreamzero} on the same GPU (\S\ref{sec:efficiency}). We validate the approach on a challenging \textit{action-free} variant of the LIBERO benchmark and on a real-world bimanual manipulator, demonstrating generalization to novel objects, novel motion primitives, and composed multi-step behaviors.

Our key contributions are as follows:
\begin{enumerate}
    \item \textbf{Action-as-VAE-patch tokenization.}
    We design an action tokenizer that reuses the video VAE patchifying convolution and matches the per-token mean and variance of its latents, adding only $1280$ parameters so the same DiT weights act on robot actions and video natively.

    \item \textbf{Three-objective joint training with a shared backbone.}
    A single pretrained video diffusion backbone is co-trained on video diffusion, joint video+action diffusion, and action-only diffusion. The joint branch uses an asymmetric attention mask and a coupled diffusion timestep across modalities.

    \item \textbf{Single-step inference.}
    Under our diffusion schedule, a single DDIM step from pure noise
    reduces inference to one forward pass on a short action token sequence, producing smooth, jerk-free action chunks. We analyze why this eliminates the high-frequency noise in world action models.

    \item \textbf{Real-world bimanual deployment.}
    We evaluate \modelname on a real bimanual manipulator, showing generalization to novel objects, novel motion primitives, and composed multi-step tasks.
\end{enumerate}

\begin{figure*}[!t]
    \centering
    \captionsetup{type=figure}
    \includegraphics[width=\textwidth]{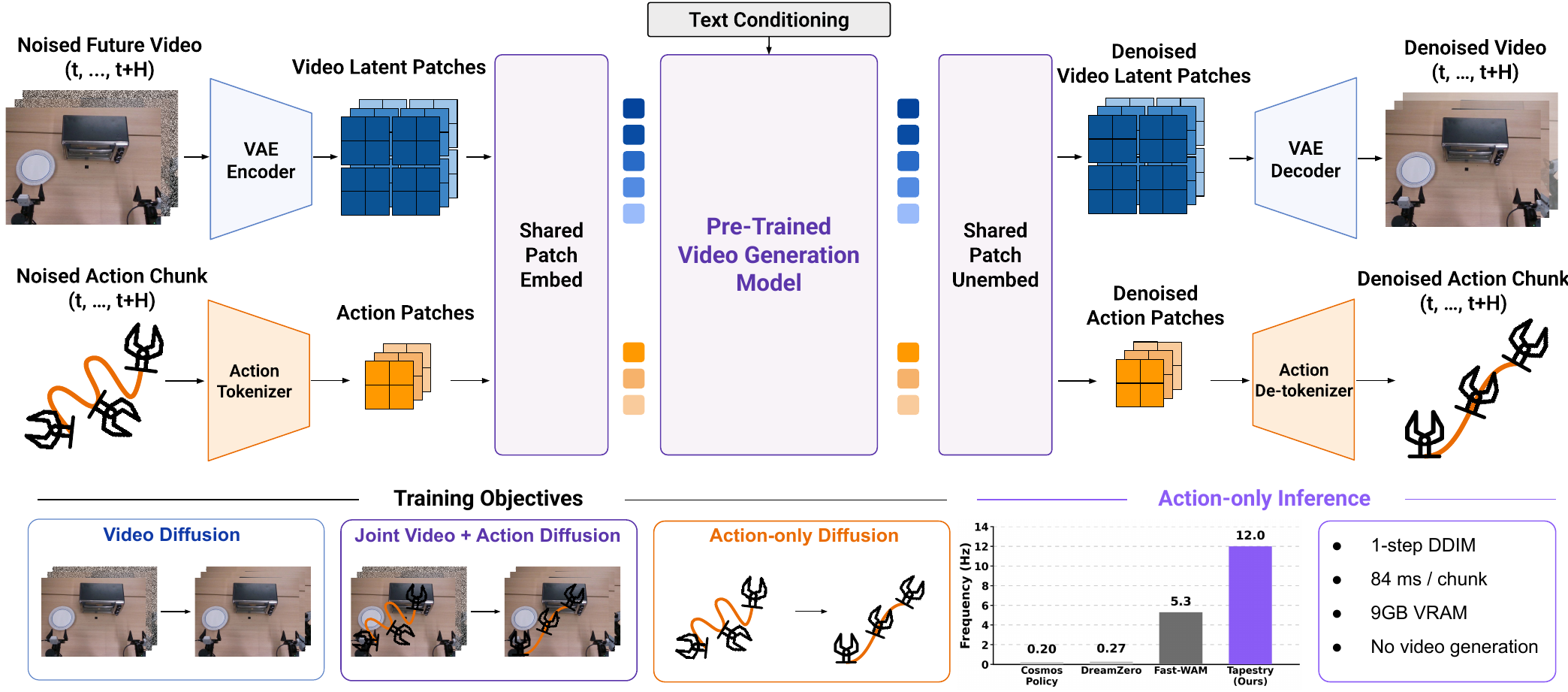}
    \caption{\modelname overview.
    The model is trained across video, joint video/action, and action-only diffusion. At deployment, it bypasses heavy video decoding to predict actions in a single step, delivering the speed and memory efficiency required for real-time robotic control.
    }
    \label{fig:banner}
\end{figure*}


\section{Method}
\paragraph{Problem Setup}
\label{sec:setup}
We assume access to two types of data: a video dataset $\mathcal{V} = \{(\mathbf{V}, g)\}$ and a robot-interaction dataset $\mathcal{R} = \{(\mathbf{V}, o, g, \mathbf{a})\}$, where $\mathbf{V}$ is a clip of pixel video, $o \in \mathcal{O}$ is the RGB observation frame at the start of $\mathbf{V}$, $g \in \mathcal{G}$ is a language goal, and $\mathbf{a} = (a_1, \dots, a_H) \in \mathbb{R}^{H \times d_a}$ is an action chunk of length $H$ over a $d_a$-dimensional action space.
Our aim is to learn a visual control policy $\pi : \mathcal{O} \times \mathcal{G} \to \mathbb{R}^{H \times d_a}$ that, at deployment, maps the current observation and language goal to an action chunk, without producing video. The policy is implemented as a single diffusion backbone trained jointly on $\mathcal{V}$ via video denoising and on $\mathcal{R}$ via joint or action-only denoising (Sec.~\ref{sec:three_objectives}), and deployed as a single denoising step on action tokens (Sec.~\ref{sec:single_step}).


\paragraph{Background and Preliminaries}
\label{sec:background}
We build on a pretrained video diffusion transformer (DiT) for text-conditioned video generation. A frozen 3D-VAE encodes a pixel video $\mathbf{V}$ into a latent tensor $\mathbf{z}^v \in \mathbb{R}^{F_v \times c \times H_z \times W_z}$ with temporal stride $4$ and spatial stride $8$ from pixel to latent dimensions, and channel count $c = 16$. Here $F_v$ counts latent frames and $H_z \times W_z$ are the spatial dimensions of each latent frame. A 2D patchifying convolution $\phi : \mathbb{R}^{p \times p \times c} \to \mathbb{R}^{d}$ with patch size $p = 2$ maps each spatial $p \times p$ patch of each latent frame to a token of hidden dimension $d$, yielding a video token sequence $\mathbf{x}^v \in \mathbb{R}^{N_v \times d}$ with sequence length $N_v = F_v \cdot H_z W_z / p^2$. The current observation frame $o$ is encoded by replicating it along the temporal axis, passing the result through the VAE, and patchifying with $\phi$ to produce observation image tokens $\mathbf{x}^o$, which together with the frozen T5-XXL~\citep{t5} text token sequence $\boldsymbol{\tau}$ form a clean conditioning prefix that is never noised. The DiT, denoted $\mathbf{v}_\theta$, is trained with the DDPM~\citep{ho2020denoising} $v$-prediction~\citep{salimans2022progressive} objective and sampled at inference with DDIM~\citep{song2020denoising}. Diffusion timesteps are $t \in \{0, 1, \dots, T\}$ where $T$ is the final diffusion timestep, and the noise schedule has \emph{zero terminal SNR} ($\bar{\alpha}_T = 0$): the fully-noised state $\mathbf{x}_T = \boldsymbol{\epsilon}$ for $\boldsymbol{\epsilon} \sim \mathcal{N}(0, I)$ carries no information about $\mathbf{x}_0$, a property we exploit at inference (\S\ref{sec:hf-investigation}). Throughout, we denote a clean token sequence by subscript $0$ and its noised version at timestep $t$ by subscript $t$.

\subsection{Action Tokenization}
\label{sec:action_tokenization}

The video patchifying convolution $\phi$ was pretrained on KL-regularized VAE patches that approximately follow $\mathcal{N}(0, I)$ statistics, and projecting raw action vectors directly into the DiT hidden space produces tokens far outside that distribution. In practice, a direct linear projection $\mathbb{R}^{d_a} \to \mathbb{R}^{d}$ that bypasses $\phi$ exhibits instability early into training, and standard remediations (gradient clipping, NaN-skipping, normalization) delay but do not eliminate the divergence. Routing actions through $\phi$ removes it and lets the same DiT weights act on action and video tokens alike. We treat each action timestep as a single VAE patch, reusing as much of the pretrained tokenization pipeline as possible. The closest analogue in the literature is Cosmos Policy~\citep{kim2026cosmospolicyfinetuningvideo}, which expresses each action chunk as 112 redundant copies tiled into the existing video latent grid.
We contrast its inference cost with ours in \S\ref{sec:efficiency}. Concretely, we map each per-step action vector $a_i \in \mathbb{R}^{d_a}$ into a DiT token via a linear projection $W \in \mathbb{R}^{(p \times p \times c) \times d_a}$, an affine-free \texttt{LayerNorm}, and the backbone's shared patchifying convolution $\phi$: $    \mathbf{x}^a_i \;=\; \phi\!\left(\mathrm{LN}\!\left(W\, a_i\right)\right) \;\in\; \mathbb{R}^{d}. $

The affine-free \texttt{LayerNorm} matches only the per-token mean and variance of the approximately unit-normal KL-regularized VAE latents that $\phi$ was pretrained on, and sharing $\phi$ rather than introducing a fresh action projection means action tokens and video tokens enter the DiT through the same learned linear map. The procedure yields one DiT token per action timestep, so the full action sequence $\mathbf{x}^a \in \mathbb{R}^{H \times d}$ has length $H$, which is typically a much shorter sequence length than the $H_z W_z / p^2$ spatial tokens per latent timestep required for video patch tokens.
Action tokens receive the backbone's fixed 3D sinusoidal positional encodings with temporal indices $0,\dots,H{-}1$ on a $1\times1$ spatial grid; in the joint mode, video and action tokens are each indexed from $0$ and then concatenated. After the DiT forward, the backbone's pretrained patch unembedding $\phi_{\text{out}}$ and the tied transpose of $W$ map each denoised action token $\mathbf{x}^{a,\text{out}}_i$ back to action space: $\hat{a}_i \;=\; W^\top \phi_{\text{out}}\!\left(\mathbf{x}^{a,\text{out}}_i\right)$.

The readout reuses $W$ and $\phi_{\text{out}}$ and adds no parameters. Together these choices keep the parameter count of the action pathway minimal: $W$ contains $p \times p \times c \cdot d_a = 2 \cdot 2 \cdot 16 \cdot 20 = 1280$ parameters (no bias), and is the \emph{only} module the action pathway adds to the pretrained backbone; $\phi$, $\phi_{\text{out}}$, and all DiT weights are inherited from pretraining and fine-tuned, and the positional encodings are fixed.

\subsection{Training Objectives}
\label{sec:three_objectives}

Each minibatch sample is trained in one or more of three modes
$m \in \{\mathrm{video}, \mathrm{joint}, \mathrm{action}\}$. Video-only samples use the
video mode; action-supervised samples use the joint mode with probability
$p_{\mathrm{joint}}=0.5$ and both the video and action-only modes otherwise. Let
$\mathbf{c}=(\boldsymbol{\tau},\mathbf{x}^o)$ denote the clean text and observation prefix,
which is shared across all modes. All modes use the same DiT
$\mathbf{v}_\theta$; only the noised tokens and loss terms differ:
\begin{equation}
\begin{aligned}
\mathcal{L}_{\mathrm{video}}
&= \lambda_v\,\mathbb{E}\!\left[
\left\|\mathbf{v}_\theta(\mathbf{x}^v_t,t,\mathbf{c})-\mathbf{v}^v_t\right\|_2^2
\right],\\
\mathcal{L}_{\mathrm{joint}}
&= \mathbb{E}\!\left[
\lambda_v\left\|\hat{\mathbf{v}}^v_\theta-\mathbf{v}^v_t\right\|_2^2
+\lambda_a\left\|\hat{\mathbf{v}}^a_\theta-\mathbf{v}^a_t\right\|_2^2
\right],\\
\mathcal{L}_{\mathrm{action}}
&= \lambda_a\,\mathbb{E}\!\left[
\left\|\mathbf{v}_\theta(\mathbf{x}^a_t,t,\mathbf{c})-\mathbf{v}^a_t\right\|_2^2
\right],
\end{aligned}
\label{eq:training_losses}
\end{equation}
where $(\lambda_v,\lambda_a)=(0.1,1)$. In the joint mode, video and action tokens are noised at
a shared timestep and passed as $(\mathbf{c},\mathbf{x}^v_t,\mathbf{x}^a_t)$, producing
predictions $(\hat{\mathbf{v}}^v_\theta,\hat{\mathbf{v}}^a_\theta)$. We use the
same AdaLN and timestep conditioning for both modalities, unlike independent
timestep formulations~\citep{zhu2025unified,kim2026cosmospolicyfinetuningvideo,ye2026dreamzero}.
An asymmetric attention mask lets action tokens attend to text, observation, and video
tokens, while preventing video tokens from attending to actions; this
preserves the pretrained video pathway while allowing actions to condition on
predicted visual dynamics.

The action-only mode drops video tokens entirely and therefore matches
deployment, where no future video is generated. This branch is empirically
critical: at the 5-epoch ablation budget, a model trained with the joint mode
alone reaches $78\%$ on LIBERO when it also generates video at inference, but
catastrophically fails when it denoises actions alone (Appendix~\ref{app:ablations}). Full
fine-tuning updates $W$ and all DiT weights, including $\phi$ and
$\phi_{\text{out}}$; the positional encodings are fixed, and the VAE and T5-XXL
encoders are frozen.

\subsection{Inference}
\label{sec:single_step}

At test time, \textit{only the action pathway is active} and matches the action-only training mode: given the current observation tokens $\mathbf{x}^o$ and text tokens $\boldsymbol{\tau}$, the DiT performs a single DDIM step from pure noise $\mathbf{x}^a_T = \boldsymbol{\epsilon}$. Under zero terminal SNR ($\bar\alpha_T = 0$) and $v$-prediction, the standard DDIM step from $t = T$ to $t = 0$ collapses to
\begin{equation}
    \hat{\mathbf{x}}^a_0 \;=\; -\,\mathbf{v}_\theta\!\left(\boldsymbol{\epsilon},\; T,\; \boldsymbol{\tau},\; \mathbf{x}^o\right), \qquad \boldsymbol{\epsilon} \sim \mathcal{N}(0, I)
    \label{eq:single_step}
\end{equation}
Detokenization then yields the predicted action chunk $\hat{\mathbf{a}}$. No video tokens are sampled and no VAE decoder is needed. Under our schedule, the one-step output approximates the conditional mean $\mathbb{E}[\mathbf{x}^a_0 \mid \boldsymbol{\tau},\, \mathbf{x}^o]$, while counterintuitively \textit{multi-step DDIM under the same backbone \emph{adds} high-frequency jitter} rather than refining the chunk. We characterize this phenomenon and identify its mechanism in Sec.~\ref{sec:hf-investigation} and Appendix~\ref{app:singlestep}.

\subsection{Implementation}
\label{sec:implementation}

We instantiate \modelname using CogVideoX-2B~\citep{cogvideox} as the pretrained backbone. The model accepts up to three camera streams, with per-view video tokens concatenated along the sequence dimension; samples with fewer views use learned mask tokens at the trailing positions to keep a fixed sequence length. Following the Large Video Planner~\citep{lvp} mixture, we train on large-scale cross-embodiment robot video (DROID~\citep{droid}, BridgeData~v2~\citep{bridgev2}, AgiBot-World~\citep{agibot}) and action-free human and internet video (Ego4D~\citep{ego4d}, Epic Kitchens~\citep{epickitchens}, SSv2~\citep{ssv2}, Panda-70M~\citep{panda70m}), both routed to the video-diffusion objective, together with in-domain bimanual teleoperation of the three target tasks plus undirected play, which supplies the only action labels (full mixture, routing, and the $20$D action space in Appendix~\ref{app:data}). Training runs on 32 H200s with 75k pretraining steps followed by 25k fine-tuning steps, and uses standard image augmentations. Full hyperparameters are in Appendix~\ref{app:hparams}.

\section{High-frequency noise and single-step inference}
\label{sec:hf-investigation}

The standard expectation in image and video diffusion is that more denoising steps yield higher-quality samples. For world action models we observe the opposite: action chunks generated with multi-step DDIM carry high-frequency jitter that degrades robot execution, while \emph{single-step} DDIM from pure noise produces smooth chunks that match the training distribution. From-scratch action diffusion approaches such as Diffusion Policy~\citep{chi2023diffusion} typically use 10--100 iterative steps and do not report this pathology, which suggests it is specific to repurposed video-generation backbones. Other video-backbone policies show traces of the same effect: DreamZero~\citep{ye2026dreamzero} reports high-frequency noise on its action chunks and applies a Savitzky-Golay filter at test time, and Cosmos Policy~\citep{kim2026cosmospolicyfinetuningvideo} averages $112$ parallel action predictions per chunk, which exhibits a smoothing effect that masks high-frequency noise. The rest of this section diagnoses the source and identifies the schedule and parameterization property under which single-step inference avoids it. Derivations, raw data, and extended figures are in Appendix~\ref{app:singlestep}.

\begin{figure}[t]
    \centering
    \begin{subfigure}[b]{0.51\linewidth}
        \centering
        \includegraphics[width=\linewidth]{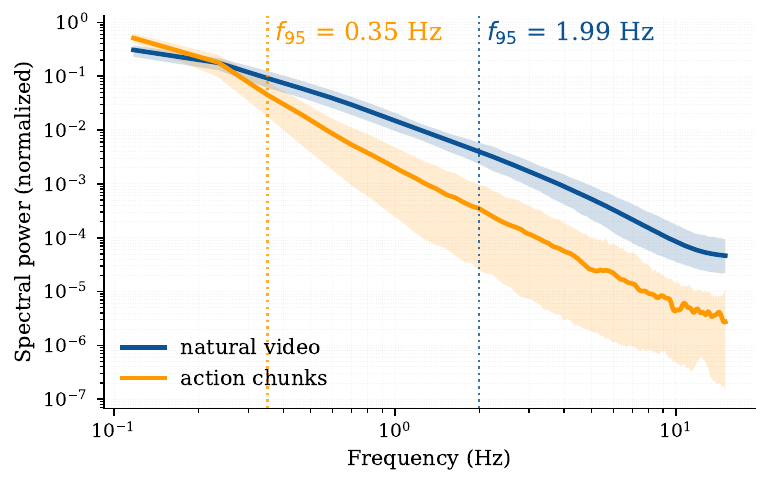}
        \caption{}
        \label{fig:hf_fft}
    \end{subfigure}
    \hfill
    \begin{subfigure}[b]{0.48\linewidth}
        \centering
        \includegraphics[width=\linewidth]{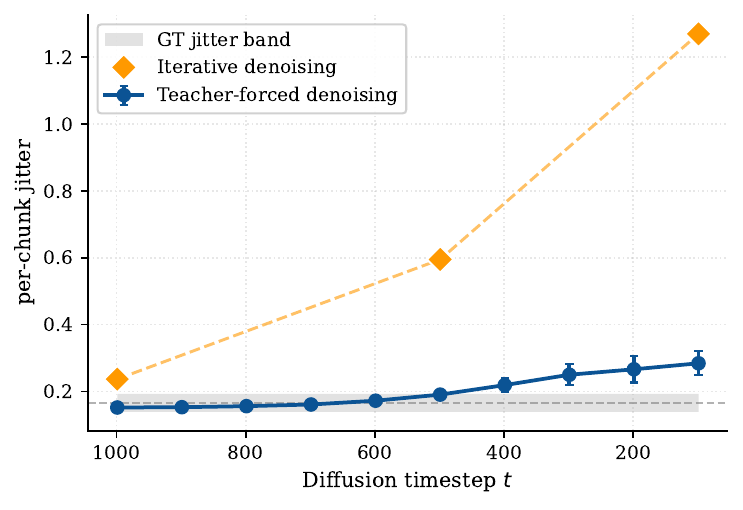}
        \caption{}
        \label{fig:hf_iter_growth}
    \end{subfigure}
    \caption{\textbf{The video-pretrained backbone has the wrong frequency prior for actions.} A backbone trained to synthesize high-frequency video detail at low $t$ keeps doing so when repurposed for action denoising, where the target distribution carries no such content. \textbf{(a) Cause: spectral mismatch in the training distributions.} $f_{95}$ (vertical dashes) is the frequency below which $95\%$ of a signal's energy sits; action chunks ($f_{95} \approx 0.35$\,Hz) carry far less high-frequency content than the natural video the backbone was pretrained on ($f_{95} \approx 2.0$\,Hz). \textbf{(b) Effect: high-frequency jitter on the action output.} Per-$t$ jitter rises as $t \to 0$ even under teacher-forcing with in-distribution noisy inputs (blue), and iterative DDIM (orange) compounds the per-step residuals into a much larger error. Single-step inference at $t = T$ sidesteps the regime entirely (\S\ref{para:hf_theory}).}
    \label{fig:hf_diagnostic_banner}
\end{figure}

We measure per-chunk RMS of the second finite difference $\|a_{i+1} - 2 a_i + a_{i-1}\|$ along the DDIM denoising trajectory on $50$ held-out validation chunks. Iterative $\hat{\mathbf{x}}_0^{(k)}$ jitter grows monotonically as $t \to 0$ (Fig.~\ref{fig:hf_iter_growth}). The 1-step output from $t = T$ matches in-distribution statistics; the same backbone diverges from them as soon as a second update is taken, and the choice of scheduler does not change this (Appendix~\ref{app:scheduler_ablation}).

We rule out four alternative explanations by direct measurement (Appendix~\ref{app:alternatives}): shared AdaLN timestep coupling, decoupled-timestep sampling (the Cosmos Policy, DreamZero, and UWM prescriptions), scheduler choice (DDIM, UniPC, zero-shift SNR), and our $v$-prediction parameterization. None distinguishes our setting from from-scratch action diffusion policies, which show no such pathology, leaving the video pretraining itself as the only remaining variable.

\paragraph{Mechanism: frequency space mismatch.}
A diffusion model implicitly learns a $t$-dependent prior over how much fine detail to synthesize at each noise level, shaped by the training distribution. Natural video carries substantial high-frequency content (textures, edges, fine motion), while action chunks do not. When a video backbone is repurposed as an action policy, this prior is incorrect in the low-$t$ regime, where the model adds fine detail to a near-clean signal: it fills in high-frequency content that does not belong in the action signal. A spectral comparison (Fig.~\ref{fig:hf_fft}; Appendix~\ref{app:fft}) shows this gap. Across chunk-horizon and long-form windows, the $f_{95}$ of natural video, the frequency below which $95\%$ of its energy lies, is $4$--$6\times$ that of action chunks. The bias is also visible in a teacher-forced sweep (Fig.~\ref{fig:hf_iter_growth}, blue) where we feed the model the \emph{correct} noised ground-truth input $\mathbf{x}_t = \sqrt{\bar{\alpha}_t}\,\mathbf{x}_0 + \sqrt{1-\bar{\alpha}_t}\,\boldsymbol{\epsilon}$ at each $t$: jitter on the resulting $\hat{\mathbf{x}}_0$ rises monotonically as $t$ decreases, reaching $1.71\times$ the GT band at $t = 99$ despite in-distribution inputs at every $t$. The bias is in the model's outputs themselves at low $t$, not in compounded off-manifold inputs. Iterative inference then compounds these per-step residuals into the orange trajectory of Fig.~\ref{fig:hf_iter_growth}.

\paragraph{Single-step inference avoids this mismatch.}
\label{para:hf_theory}
Single-step inference from pure noise sidesteps the mis-tuned prior by querying the model only at $t = T$, where the target is structurally noise-independent. The collapse to $\hat{\mathbf{x}}_0 = -\mathbf{v}_\theta(\boldsymbol{\epsilon}, T, \mathbf{c})$ in Eq.~\ref{eq:single_step} holds because at $\bar{\alpha}_T = 0$ the $v$-target reduces to $-\mathbf{x}_0$, becoming \emph{$\boldsymbol{\epsilon}$-independent} by construction (derivation in Appendix~\ref{app:derivation}); we measure $\mathrm{Var}_{\boldsymbol{\epsilon}}\,\mathbf{v}_\theta = 3.5{\times}10^{-6}$ at $t = T$, five orders of magnitude below its value at $t = 99$. An $\boldsymbol{\epsilon}$-prediction parameterization, a non-zero terminal-SNR schedule, or flow-matching at $\sigma_{\max} = 1$ each keeps the target $\boldsymbol{\epsilon}$-dependent at the terminus and restores the jitter (Appendix~\ref{app:wan_comparison}).

\section{Experiments}
\label{sec:experiments}

We evaluate \modelname across four axes:
(i)~learning-from-video task generalization on a non-standard variant of the LIBERO simulation benchmark (\S\ref{sec:libero}), in which the model receives \emph{no action labels} for the evaluation suites and must transfer policies from videos alone;
(ii)~real-world bimanual manipulation (\S\ref{sec:realworld}), measuring in-distribution success and generalization to novel object placements, motion primitives, and object compositions;
(iii)~ablations of each architectural and training-recipe choice (\S\ref{sec:ablations});
and~(iv)~inference efficiency relative to other joint video--action policies  (\S\ref{sec:efficiency}).

\subsection{LIBERO task generalization}
\label{sec:libero}

\paragraph{Setup.}
LIBERO~\citep{liu2023libero} comprises five task suites: LIBERO-10, 90, Object, Spatial, and Goal.
We adopt the \emph{learning-from-video} protocol of AMPLIFY~\citep{collins2025amplifyactionlessmotionpriors}: action labels are provided \emph{only} for LIBERO-90, and the four evaluation suites (LIBERO-10, Object, Spatial, Goal) are seen as video alone.
Every method must therefore bridge from eval-task videos to actions, so absolute success rates are not comparable to the standard fully-supervised LIBERO leaderboard.

\paragraph{Baselines.}
Behavior-cloning baselines that train on LIBERO-90 actions only and ignore videos: QueST~\citep{mete_quest_2024}, BAKU~\citep{haldar2024baku}, and Diffusion Policy~\citep{chi2023diffusion}.
Learning-from-video methods that also leverage the held-out video data: AMPLIFY~\citep{collins2025amplifyactionlessmotionpriors}, Cosmos Policy~\citep{kim2026cosmospolicyfinetuningvideo}, and DreamZero~\citep{ye2026dreamzero}. Cosmos Policy is trained for 190k gradient steps and DreamZero is trained for 100k gradient steps until convergence, and all models are trained on the same data as \modelname{}.
We additionally report an ablation of our method trained from scratch (without video pretraining) to isolate the benefit of the pretrained video backbone.

\paragraph{Results.}
Table~\ref{tab:libero} reports held-out success on the four evaluation suites. \modelname{} reaches $76\%$, $6$ points above the strongest learning-from-video baselines (Cosmos Policy and DreamZero, both $70\%$) and $15$ above AMPLIFY. The margin is uneven: Cosmos Policy and DreamZero lead on LIBERO-10 and DreamZero is slightly ahead on Goal, while \modelname{} wins decisively on Object and Spatial, the suites that most reward transferring object and geometric priors from video. Behavior-cloning baselines that ignore the held-out video collapse (Diffusion Policy, QueST, BAKU): LIBERO-90 actions alone do not transfer. A from-scratch ablation drops to $58\%$, isolating $18$ points to the pretrained backbone.

\begin{table}[t]
\centering
\caption{\textbf{LIBERO task generalization.} Success rates (\%) on four held-out LIBERO suites. The model is trained with action labels only from LIBERO-90 and video data from all suites until convergence. Each suite is evaluated over $100$ rollouts.}
\label{tab:libero}
\begin{tabular}{@{}lccccc@{}}
\toprule
Method & LIBERO-10 & Object & Spatial & Goal & Avg \\
\midrule
\multicolumn{6}{@{}l}{\emph{Behavior-cloning baselines (LIBERO-90 action labels only)}} \\
QueST~\citep{mete_quest_2024}                            & 7  & 0  & 1  & 1  & 2 \\
\rowcolor[HTML]{EFEFEF}
BAKU~\citep{haldar2024baku}                              & 6  & 0  & 0  & 0  & 2 \\
Diffusion Policy~\citep{chi2023diffusion}                & 0  & 0  & 0  & 0  & 0 \\
\midrule
\multicolumn{6}{@{}l}{\emph{Learning-from-video methods (LIBERO-90 actions + held-out videos)}} \\
AMPLIFY~\citep{collins2025amplifyactionlessmotionpriors} & 52 & 80 & 69 & 41 & 61 \\
\rowcolor[HTML]{EFEFEF}
Cosmos Policy~\citep{kim2026cosmospolicyfinetuningvideo} & \textbf{70} & 88 & 79 & 44 & 70 \\
DreamZero~\citep{ye2026dreamzero}   & 60  & 84  & 68  & \textbf{66}  & 70 \\
\modelname (from scratch)  & 40 & 80 & 75 & 35 & 58 \\
\rowcolor[HTML]{EFEFEF}
\modelname (pretrained, ours) & 55 & \textbf{95} & \textbf{87} & 65 & \textbf{76} \\
\bottomrule
\end{tabular}
\end{table}


\subsection{Real-world bimanual manipulation}
\label{sec:realworld}

\vspace*{-3pt}\paragraph{Setup.}
We deploy \modelname on a 20-DoF bimanual YAM platform equipped with wrist and third-person cameras (Fig.~\ref{fig:hardware_setup}).
Training data are in-house teleoperation demonstrations of three target tasks: \textit{Toolbox} (insert a screwdriver into a toolbox), \textit{Oven} (open the oven and place the croissant on the plate),
and \textit{Corn} (place a corn cob in a bowl). For each task, we run 10 rollouts from a randomized initial state distribution and report success rates.

To probe generalization, we evaluate three out-of-distribution conditions (Fig.~\ref{fig:ood_setup}), each isolating a distinct transfer axis.
\textbf{(1) Novel object placement:} the croissant is placed on a \emph{transparent glass oven door} rather than inside the oven, shifting only the initial state.
\textbf{(2) Novel motion primitive:} the robot pours an unseen golf ball from an unseen bowl, a motion primitive appearing only sparsely in play data.
\textbf{(3) Novel object composition:} the robot places a Lego brick on a wooden boomerang, neither seen in any robot demonstration, isolating the contribution of the cross-embodiment video data.

\vspace{-3pt} \paragraph{Baselines.}
We compare against a Diffusion Policy~\citep{chi2023diffusion} DiT variant trained from scratch on our in-domain robot data. On the out-of-distribution conditions, we additionally compare against DreamZero~\citep{ye2026dreamzero}, trained with its default hyperparameters on the same data mixture as \modelname{} (full-finetuning Wan2.2-TI2V-5B for $25$k pretraining and $25$k fine-tuning steps; Appendix~\ref{app:hparams}).
We also include two ablations of \modelname{} with components removed: \emph{w/o large-scale video co-training} trains only on the in-domain robot videos and actions, and \emph{w/o video generation objective} additionally drops the video diffusion objective entirely.

\vspace{-3pt} \paragraph{Results.}
On in-distribution success (Table~\ref{tab:realworld_indist}), \modelname{} reaches $93 \pm 5\%$ against $50 \pm 9\%$ for the from-scratch BC baseline ($p<0.001$); removing large-scale video co-training gives $67\%$, additionally dropping the video generation objective gives $63\%$, and only the full recipe is strong on every task.
The gap widens out of distribution (Table~\ref{tab:realworld_gen}): DreamZero, despite inheriting motion priors from video generation, solves only the pour task and scores $0/5$ on placement and composition, while the from-scratch and ablated variants collapse to $0\%$. \modelname{} reaches $67 \pm 12\%$ and is the only method to complete the novel object composition, which most depends on the long tail of objects in the cross-embodiment video data. With $5$ rollouts per condition, the margin over DreamZero ($27 \pm 11\%$, $p=0.07$) is not significant, while the margin over the other methods ($0 \pm 0\%$, $p<0.001$) is. A symmetric-target probe with two identical croissants and no disambiguating language succeeded in $8/10$ rollouts, matching an $8/10$ single-croissant control, and every reach committed to a single croissant (Appendix~\ref{app:symmetric_probe}).
Per-subtask counts (Appendix~\ref{sec:realworld-subtasks}) show ablations and BC typically clear the approach but fail terminal contact, e.g.\ BC opens the oven on all $10$ rollouts but places the croissant on only $4$.

\begin{table}[!t]
\centering
\caption{\textbf{Real-world in-distribution success.} Cumulative success counts on the three trained tasks, 10 rollouts per task.}
\label{tab:realworld_indist}
\begin{tabular}{@{}lc!{\color{black!25}\vrule}c!{\color{black!25}\vrule}c!{\color{black!25}\vrule}c@{}}
\toprule
Method & Toolbox & Oven & Corn & Total \\
\midrule
Diffusion Policy~\citep{chi2023diffusion} & 3/10 & 4/10 & 8/10 & 15/30 \\
\rowcolor[HTML]{EFEFEF}
\modelname w/o video generation objective & 5/10 & \textbf{9/10} & 5/10 & 19/30 \\
\modelname w/o large-scale video co-training & 4/10 & 7/10 & 9/10 & 20/30 \\
\rowcolor[HTML]{EFEFEF}
\modelname (ours) & \textbf{9/10} & \textbf{9/10} & \textbf{10/10} & \textbf{28/30} \\
\bottomrule
\end{tabular}
\end{table}


\begin{table}[!t]
\vspace{-7mm}
\centering
\small
\caption{\textbf{Real-world out-of-distribution generalization with partial and full success rates.} Three OOD conditions probe novel object placement (croissant on transparent oven door), motion primitive (pour an unseen golf ball out of an unseen bowl), and object composition (place an unseen yellow Lego brick on an unseen wooden boomerang). Tasks additionally include unseen distractor objects in the scene and perturbed initial states. \emph{Partial} credits the rollout for clearing the first subtask of each task (e.g., grasping the target object); \emph{Full} requires the rollout to complete the entire task. Each condition is evaluated over $5$ rollouts.}
\label{tab:realworld_gen}
\resizebox{\linewidth}{!}{
\begin{tabular}{@{}lcc!{\color{black!25}\vrule}cc!{\color{black!25}\vrule}cc!{\color{black!25}\vrule}c@{}}
\toprule
Method & \multicolumn{2}{c!{\color{black!25}\vrule}}{\makecell{Croissant on \\transparent door}} & \multicolumn{2}{c!{\color{black!25}\vrule}}{\makecell{Pour golf ball \\out of bowl}} & \multicolumn{2}{c!{\color{black!25}\vrule}}{\makecell{Lego brick on \\boomerang}} & Total \\
\cmidrule(lr){2-3} \cmidrule(lr){4-5} \cmidrule(lr){6-7}
 & Partial & Full & Partial & Full & Partial & Full & \\
\midrule
Diffusion Policy~\citep{chi2023diffusion} & 0/5 & 0/5 & 1/5 & 0/5 & 1/5 & 0/5 & 0/15 \\
\rowcolor[HTML]{EFEFEF}
DreamZero~\citep{ye2026dreamzero} & 0/5 & 0/5 & \textbf{5/5} & \textbf{4/5} & 1/5 & 0/5 & 4/15 \\

\modelname w/o video generation objective & 0/5 & 0/5 & 0/5 & 0/5 & 4/5 & 0/5 & 0/15 \\
\modelname w/o large-scale video co-training & 0/5 & 0/5 & 2/5 & 0/5 & 3/5 & 0/5 & 0/15 \\
\rowcolor[HTML]{EFEFEF}
\modelname (ours) & \textbf{3/5} & \textbf{1/5} & \textbf{5/5} & \textbf{4/5} & \textbf{5/5} & \textbf{5/5} & \textbf{10/15} \\
\bottomrule
\end{tabular}
}
\end{table}

\subsection{Inference efficiency}
\label{sec:efficiency}

A direct consequence of treating each action timestep as a single VAE-patch token (rather than as a full latent frame or a long-horizon video sequence) is that the inference-time token budget collapses to a small action chunk.
Combined with single-step DDIM inference (\S\ref{sec:hf-investigation}), this yields substantially lower wall-clock cost than other joint video--action policies.
On a single consumer-grade RTX~5090, \modelname produces a 16-action chunk in $84$\,ms with $9$\,GB of GPU memory. On the same GPU, DreamZero, a 5B joint video--action policy, takes ${\sim}20$\,s per chunk with its released code and $3.7$\,s after we add text-embedding caching and activation checkpointing (neither in the release), ${\sim}44\times$ slower than \modelname. The gap comes from denoising $16$ action tokens in one forward pass, where DreamZero denoises video and action tokens together over $4$ forward passes with a larger backbone (5B versus 2B). Training compute is comparable to DreamZero's (Appendix~\ref{app:hparams}).

\subsection{Ablations}
\label{sec:ablations}

In order to understand the core mechanisms behind our method, we ablate the key design decisions of \modelname{} on LIBERO at a fixed 5-epoch budget (Table~\ref{tab:ablations_libero}, Appendix~\ref{app:ablations}), where the full recipe reaches $58\%$ ($76\%$ at convergence). Bypassing $\phi$ with a direct linear projection diverges, and removing the affine-free LayerNorm drops success to $33\%$. Separate video and action objectives on the same data, initialization, and hyperparameters reach $36\%$; dropping the video loss entirely gives $0\%$, and training from scratch gives $13\%$. The joint video$+$action variant ($78\%$) generates video at inference (\S\ref{sec:three_objectives}).

\section{Related Work}

Prior work on video-pretrained policies divides along two questions: where the pretrained backbone $\theta$ acts, and what the model emits at deployment.
One family generates a full predicted video $\hat{\mathbf{V}}$ from $(\boldsymbol{\tau}, \mathbf{x}^o)$ and recovers actions through a separate inverse-dynamics module; UniPi~\citep{du2023learning}, AVDC~\citep{ko2024learning}, and DreamGen~\citep{jang2025dreamgen} are representative. Inference cost scales with the full video token count $N_v$ rather than the action horizon $H$, and pseudo-action fidelity is bounded by the worse of video quality and IDM accuracy.
A second family avoids generating video entirely. The backbone $\theta$ runs a single noisy forward pass, an intermediate feature $h_\ell$ is extracted, and a from-scratch policy head $g_\psi$ maps $h_\ell$ to actions; VidMan~\citep{wen2024vidman}, VPP~\citep{hu2024video}, mimic-video~\citep{zhong2025mimicvideo}, and Video Policy~\citep{liang2025video} take this route. Here $\theta$ influences actions only indirectly through $h_\ell$, and the action tokenization is decoupled from the backbone's pretrained input distribution.

The closest work to ours routes both video and action tokens through the same backbone $\mathbf{v}_\theta$. UWM~\citep{zhu2025unified} draws independent timesteps $(t^v, t^a)$ but trains from scratch, so $\theta$ inherits no pretrained video prior. Cosmos Policy~\citep{kim2026cosmospolicyfinetuningvideo} encodes each action chunk as $112$ replicated copies into the video latent grid,
and runs at ${\sim}0.2$\,Hz. DreamZero~\citep{ye2026dreamzero} jointly denoises video and action on a Wan~2.1 backbone but still emits the full video sequence at test time. Fast-WAM~\citep{yuan2026fastwamworldactionmodels} keeps a fine-tuned video DiT but routes actions through a separate from-scratch action expert, so the video backbone $\theta$ does not directly process $\mathbf{x}^a$. GigaWorld-Policy~\citep{ni2026gigaworld} makes future video optional at inference but embeds actions with new linear projections into the DiT hidden dimension and decodes them with a new flow-matching head sampled iteratively. Other unified architectures~\citep{li2025unified, bi2025motus, won2025dual} concatenate both modalities into one transformer sequence and pay for the full video token block at deployment. \modelname{} differs in two places. First, actions enter and leave through the backbone's own patchifier $\phi$ and unembedding $\phi_{\text{out}}$, so $\theta$ receives action tokens whose per-token statistics match its pretraining inputs, and the action pathway adds only $W$. Second, deployment drops the video token block and takes a single denoising step, which we show is well-posed under zero-terminal-SNR $v$-prediction (\S\ref{sec:hf-investigation}).
Extended related work in Appendix~\ref{app:related-work} expands the per-method comparison.

\section{Conclusion}

We presented \modelname, a method for repurposing a pretrained video diffusion transformer into a robotic policy without performing video generation at inference. The recipe combines three ingredients: an action tokenizer that reuses the video VAE's patchifying convolution and matches the per-token mean and variance of its latents; a three-objective joint training procedure on a single shared backbone; and single-step DDIM inference, which under our zero-terminal-SNR $v$-prediction schedule reduces to a single forward pass from pure noise. Together these choices yield a policy that runs on a single consumer-grade RTX~5090, ${\sim}44\times$ faster than an optimized DreamZero on the same GPU, while improving on the strongest published baseline on LIBERO held-out task generalization.

\paragraph{Limitations.}
Single-step inference returns the conditional mean of the action chunk, one deterministic prediction per observation. We observed no averaging in our evaluations, including a symmetric two-target probe (Appendices~\ref{app:cond_mean_limitation}--\ref{app:symmetric_probe}), but tasks with multimodal or highly stochastic actions may need iterative refinement or one of the remedies discussed there. The single-step result relies on the zero-terminal-SNR $v$-prediction schedule of our backbone; on a rectified-flow backbone (Wan), replacing the native objective was unstable in initial experiments, and whether single-step inference carries over remains open (Appendix~\ref{app:wan_comparison}). $W$ is sized to our $20$D bimanual action space, and per-embodiment action heads are unexplored. 
\section*{Acknowledgements}
We thank Namra Patel, Dennis Anthony, Nhi Nguyen, and Keerthi Kaashyap for helpful discussions, as well as their assistance with infrastructure and data collection contributing to this work. We additionally thank Walker Byrnes for his assistance with experiments for the rebuttal.


\newpage
\bibliography{main}
\newpage
\input{appendix.tex}

\end{document}

%% file: preamble.tex
\usepackage[utf8]{inputenc} 
\usepackage[T1]{fontenc}    
\usepackage{fonttable}
\usepackage[english]{babel} 
\usepackage[protrusion=true,expansion=true]{microtype}
\usepackage{lmodern}
\usepackage{comment}
\usepackage{cancel}
\usepackage{array}
\usepackage{colortbl}
\usepackage{csquotes}
\usepackage{listings}
\usepackage{nameref}
\usepackage{paralist}
\usepackage{xspace}
\usepackage{setspace}
\usepackage{array}
\usepackage{makeidx}
\usepackage{pdfpages}
\usepackage{times}

\usepackage{soul}

\usepackage{amsmath,amssymb} 
\usepackage{amsfonts}       
\usepackage{mathrsfs} 
\usepackage{mathtools}
\usepackage{array} 
\usepackage{nicefrac}       
\usepackage{textcomp}
\usepackage{bm} 
\usepackage{pifont} 

\usepackage{graphicx}
\usepackage{wrapfig}
\usepackage{adjustbox} 
\usepackage{epsfig}
\pdfoutput=1
\graphicspath{{./figures/}}
\DeclareGraphicsExtensions{.pdf,.jpeg,.png,.jpg}

\usepackage{float}
\usepackage{placeins}
\usepackage{subcaption}
\usepackage[font=small,labelfont=bf]{caption}
\usepackage{lscape}      

\usepackage{tikz}  
\usepackage{makecell}
\usepackage{overpic}
\newsavebox{\subfigbox}

\usepackage{algorithm}
\usepackage[noend]{algpseudocode}

\usepackage{array} 
\usepackage{tabularx}
\usepackage{multirow}
\usepackage{multicol}
\usepackage{booktabs}   
\usepackage{tablefootnote}
\usepackage{threeparttable}

\usepackage{paralist}
\usepackage{enumitem}
\setlist[itemize]{noitemsep,leftmargin=*,topsep=0in}
\setlist[enumerate]{noitemsep,leftmargin=*,topsep=0in}

\usepackage{url}            
\usepackage[table]{xcolor}
\usepackage{color,colortbl} 
\usepackage{soul}

\PassOptionsToPackage{capitalize, noabbrev}{cleveref}

\usepackage{hyperref}

\usepackage{titlesec}
\titlespacing{\section}{0pt}{0.15\baselineskip}{0.1\baselineskip}
\titlespacing{\subsection}{0pt}{0.1\baselineskip}{0.08\baselineskip}
\titlespacing{\subsubsection}{0pt}{0.03\baselineskip}{0.02\baselineskip}

\renewcommand{\paragraph}[1]{\noindent\textbf{#1}\xspace}

\usepackage{blindtext}
\usepackage{bbm}

\usepackage{lipsum}

\definecolor{color1}{rgb}{0.6, 0.4, 0.05}
\definecolor{color2}{rgb}{0.0, 0.7, 0.7}
\definecolor{color3}{rgb}{0.35, 0.75, 0.0}
\definecolor{color4}{rgb}{0.4, 0.8, 0.0}
\definecolor{color5}{rgb}{0.5, 0.0, 0.5}
\definecolor{color6}{rgb}{0.99, 0.84, 0.69}
\definecolor{color7}{rgb}{0.8, 0.3, 0.3}
\definecolor{color8}{rgb}{0.2, 0.4, 0.8}
\definecolor{color9}{rgb}{0.6, 0.2, 0.6}
\definecolor{revision_color}{rgb}{1,0,0}



%% file: appendix.tex
\appendix

\section{Extended related work}
\label{app:related-work}

We compare to prior approaches along three axes: (i) what the model emits at deployment (full $\hat{\mathbf{V}}$, an internal feature, or only $\hat{\mathbf{a}}$); (ii) whether the pretrained DiT weights $\theta$ act directly on action tokens $\mathbf{x}^a$ or only via a separately trained head; and (iii) how actions are tokenized relative to the backbone's pretrained input distribution.

\paragraph{Video Generation with Inverse Dynamics.}
This body of work generates a full video rollout $\hat{\mathbf{V}}$ from $(\boldsymbol{\tau}, \mathbf{x}^o)$ and recovers actions through a separate inverse-dynamics module $\pi_{\text{IDM}}(\hat{\mathbf{V}}) \to \hat{\mathbf{a}}$. UniPi~\citep{du2023learning} pairs a text-to-video model with an IDM trained on action-labeled rollouts; AVDC~\citep{ko2024learning} replaces the IDM with optical-flow correspondences between adjacent generated frames; DreamGen~\citep{jang2025dreamgen} fine-tunes image-to-video backbones on robot embodiment data and recovers pseudo-actions via an IDM or via the LAPA~\citep{ye2024latent} latent-action model. Inference cost scales as $N_v \cdot K$ rather than $H$, and pseudo-action fidelity is bounded by both video quality and IDM accuracy.

\paragraph{Conditioning on Video-Model Latents.}
Conditioning on video model latents is a way to avoid generating $\hat{\mathbf{V}}$ at test time: $\mathbf{v}_\theta$ runs at a single noise level $t^\star$, intermediate features $h_\ell$ are extracted, and a separately trained head $g_\psi(h_\ell, \boldsymbol{\tau}, \mathbf{x}^o) \to \hat{\mathbf{a}}$ produces the action. VidMan~\citep{wen2024vidman} attaches an adapter to a robot-pretrained video diffusion model and reuses its outputs as inverse-dynamics features; VPP~\citep{hu2024video} conditions a diffusion-policy head on \emph{predictive visual representations} from a single noisy forward at $t^\star$; mimic-video~\citep{zhong2025mimicvideo} finds that \emph{partially denoised} latents at intermediate $t^\star$ best balance captured dynamics with retained signal, with policy quality scaling directly with the underlying video model; Video Policy~\citep{liang2025video} runs video and action branches in parallel and shares only the conditioning encoder. In this family $g_\psi$ is trained from scratch and $\theta$ acts on $\mathbf{x}^a$ only indirectly via $h_\ell$, so action tokenization is decoupled from the backbone's pretrained input distribution.

\paragraph{Video Generation Models as a Shared Action Backbone.}
This category of work routes both video and action tokens through a single backbone $\mathbf{v}_\theta$. UWM~\citep{zhu2025unified} draws independent timesteps $(t^v, t^a)$, allowing the same network to represent policies, forward dynamics, and inverse dynamics by routing one timestep to maximum noise; Appendix~\ref{app:alternatives} tests decoupled-timestep variants against our coupled $t^v = t^a$ design. UWM trains from scratch, however, so $\theta$ inherits no pretrained video prior. Cosmos Policy~\citep{kim2026cosmospolicyfinetuningvideo} fine-tunes Cosmos-Predict2 to jointly emit actions, future frames, and value estimates, replicating each action chunk as $112$ copies into the $H_z W_z / p^2$ video latent grid, and inference runs at ${\sim}0.2$\,Hz. DreamZero~\citep{ye2026dreamzero}, the closest concurrent work, jointly denoises $(\mathbf{x}^v_t, \mathbf{x}^a_t)$ on a fine-tuned Wan~2.1 backbone and shows cross-embodiment transfer from human video, but inference still emits the full $N_v$-token video sequence. Fast-WAM~\citep{yuan2026fastwamworldactionmodels} fine-tunes the video DiT but routes action prediction through a from-scratch action expert $\mathbf{v}_\psi$, so $\theta$ acts on $\mathbf{x}^v$ but never directly on $\mathbf{x}^a$. GigaWorld-Policy~\citep{ni2026gigaworld} makes future video optional at inference but embeds actions with new linear projections into the DiT hidden dimension and decodes them with a new flow-matching head sampled iteratively, so its action tokenization is decoupled from the backbone's pretrained input distribution. Other unified architectures~\citep{li2025unified, bi2025motus} concatenate both modalities into one transformer sequence, paying for the full video token block at deployment; DUST~\citep{won2025dual} mitigates this through a dual-stream design with asynchronous sampling. \modelname{} differs by routing actions through the same patchifying convolution $\phi$ used for video patches, $\mathbf{x}^a_i = \phi(\mathrm{LN}(W a_i))$ with $W \in \mathbb{R}^{p^2 c \times d_a}$ as the only module added for actions ($1280$ new parameters; \S\ref{sec:action_tokenization}), and dropping $\mathbf{x}^v$ from the deployment forward via action-only training (\S\ref{sec:three_objectives}), which collapses the inference token budget from $N_v \sim 10^3$ to $H = 16$ while still passing $\mathbf{x}^a$ through pretrained $\theta$.


\section{Per-subtask real-world results}
\label{sec:realworld-subtasks}

Tables~\ref{tab:realworld_subtasks_indist} and~\ref{tab:realworld_subtasks_ood} report cumulative subtask success counts for the in-distribution and out-of-distribution evaluations of \S\ref{sec:realworld}. \emph{Cumulative} means that a rollout failing at subtask $k$ counts as a failure on every subsequent subtask in the same task; a successful rollout credits all subtasks up to and including the final one. In-distribution cells are out of 10 rollouts and the per-task final-subtask counts agree with Table~\ref{tab:realworld_indist}; OOD cells are out of 5 rollouts and the per-condition final-subtask counts agree with Table~\ref{tab:realworld_gen}.

\begin{table}[h]
\centering
\small
\setlength{\tabcolsep}{4pt}
\caption{\textbf{Per-subtask cumulative success counts, in-distribution} (out of 10 rollouts per cell). A failure at one subtask propagates as a failure for all subsequent subtasks in the same rollout.}
\label{tab:realworld_subtasks_indist}
\begin{tabular}{@{}llcccc@{}}
\toprule
Task & Subtask & BC & \makecell{w/o lg-scale\\video} & \makecell{w/o video\\gen.\ obj.} & Ours \\
\midrule
Toolbox & Pick screwdriver  & 6  & \textbf{10} & 6 & \textbf{10} \\
        & Place screwdriver & 3  & 4  & 5 & \textbf{9} \\
\midrule
Oven    & Open oven         & \textbf{10} & 9  & 9 & \textbf{10} \\
        & Grasp croissant   & 6  & 7  & \textbf{9} & \textbf{9} \\
        & Place croissant   & 4  & 7  & \textbf{9} & \textbf{9} \\
\midrule
Corn    & Grab corn         & 8  & 9  & 5 & \textbf{10} \\
        & Place corn        & 8  & 9  & 5 & \textbf{10} \\
\bottomrule
\end{tabular}
\end{table}

\begin{figure}[h]
\centering
\includegraphics[width=\linewidth]{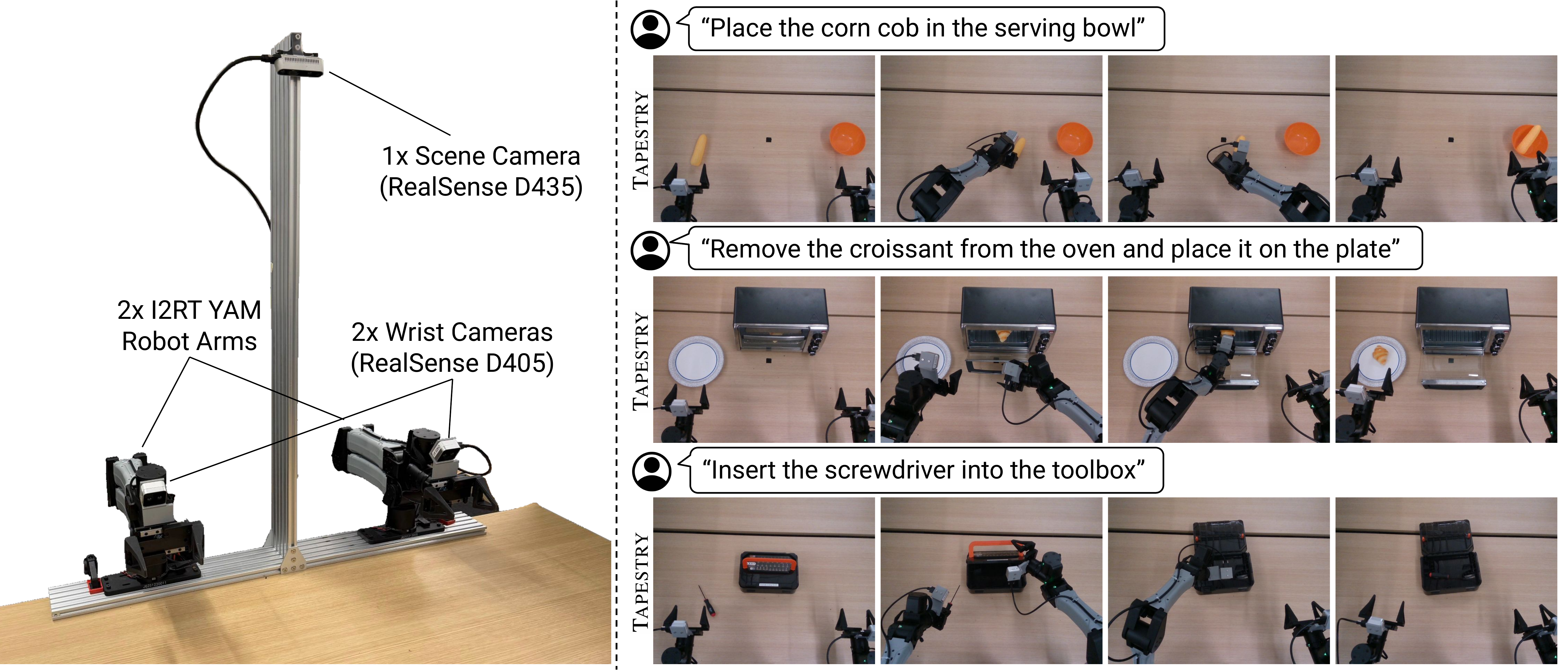}
\caption{Real-world bimanual setup: a 20-DoF bimanual YAM platform with wrist and third-person cameras, used for the experiments in \S\ref{sec:realworld}.}
\label{fig:hardware_setup}
\end{figure}

\begin{figure}[h]
\centering
\includegraphics[width=\linewidth]{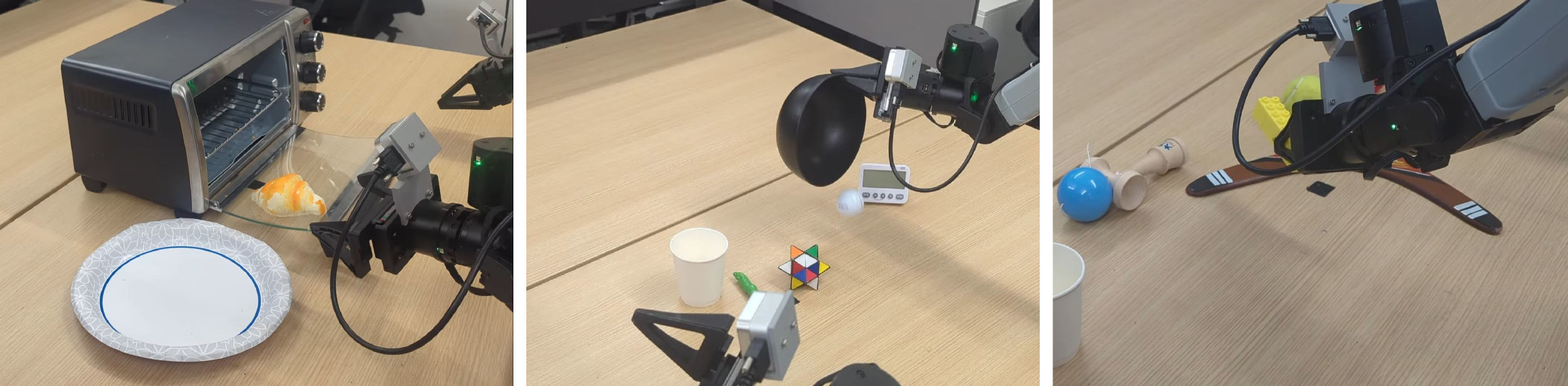}
\caption{\textbf{Out-of-distribution evaluation scenes.} Rollout snapshots for each of the three OOD conditions. \emph{Left:} novel object placement: croissant on the transparent glass oven door rather than inside the oven. \emph{Center:} novel motion primitive: pour an unseen golf ball from an unseen bowl. \emph{Right:} novel object composition: place an unseen yellow Lego brick on an unseen wooden boomerang.}
\label{fig:ood_setup}
\end{figure}

\begin{table}[h]
\centering
\small
\setlength{\tabcolsep}{4pt}
\caption{\textbf{Per-subtask cumulative success counts, out-of-distribution} (out of 5 rollouts per cell). Conditions: novel object placement (croissant on transparent oven door), novel motion primitive (pour golf ball out of bowl), novel object composition (place yellow Lego brick on a wooden boomerang).}
\label{tab:realworld_subtasks_ood}
\begin{tabular}{@{}llcccc@{}}
\toprule
Condition & Subtask & BC & \makecell{w/o lg-scale\\video} & \makecell{w/o video\\gen.\ obj.} & Ours \\
\midrule
Croissant on door  & Pick up croissant         & 0 & 0 & 0 & \textbf{3} \\
                   & Place croissant on plate  & 0 & 0 & 0 & \textbf{1} \\
\midrule
Pour golf ball     & Pick up bowl              & 1 & 2 & 0 & \textbf{5} \\
                   & Pour golf ball out        & 0 & 0 & 0 & \textbf{4} \\
\midrule
Lego on boomerang  & Pick up Lego              & 1 & 3 & 4 & \textbf{5} \\
                   & Place Lego on boomerang   & 0 & 0 & 0 & \textbf{5} \\
\bottomrule
\end{tabular}
\end{table}

\section{Ablation details}
\label{app:ablations}

We ablate the key design decisions of \modelname{} on the LIBERO learning-from-video protocol (\S\ref{sec:libero}). Each row varies a single component while holding all other components at the full-recipe setting. The joint video$+$action row reaches $78\%$ with video generation at inference; trained without the action-only mode, it fails under action-only inference. The separate-objectives and full-recipe rows ($36\%$ and $58\%$) share data, initialization, and hyperparameters and differ only in objective: separate per-modality denoising, or the joint objective on half of the action-labeled samples and the separate video and action-only objectives on the rest.

\begin{table}[h]
\centering
\caption{\textbf{Ablations on LIBERO task generalization.} Average held-out success rate (\%) over 10 rollouts per task ($100$ per suite). All variants trained for 5 epochs and share the LIBERO-90 action data and the evaluation protocol of Table~\ref{tab:libero}. At this budget the full recipe reaches $58\%$ and the from-scratch variant $13\%$; at convergence they reach $76\%$ and $58\%$ (Table~\ref{tab:libero}). \emph{Inference}: whether the evaluated model denoises actions alone or generates video jointly with actions.}
\label{tab:ablations_libero}
\begin{tabular}{@{}lcc@{}}
\toprule
Variant & Inference & Avg Success Rate \\
\midrule
\multicolumn{3}{@{}l}{\emph{Pretraining strategy}} \\
\quad From scratch (no pretrained CogVideoX init) & action & 13 \\
\quad Pretrained init, action-only fine-tuning   & action & 0 \\
\quad Pretrained init, separate video and action objectives & action & 36 \\
\midrule
\multicolumn{3}{@{}l}{\emph{Diffusion objective composition}} \\
\quad Action-only training (no video loss)             & action & 0 \\
\quad Joint video$+$action (no action-only branch) & video$+$action & 78 \\
\quad Joint $+$ action-only mix  & action & 58 \\
\midrule
\multicolumn{3}{@{}l}{\emph{Action tokenization}} \\
\quad Direct linear projection (bypass shared conv) & -- & NaN (diverges) \\
\quad Without affine-free LayerNorm                 & action & 33 \\
\midrule
\modelname{} & action & \textbf{58} \\
\bottomrule
\end{tabular}
\end{table}

\section{Hyperparameters and training details}
\label{app:hparams}

Table~\ref{tab:hparams} lists the full set of training hyperparameters. Image augmentations (random resize crop, color jitter, Gaussian blur) are applied to the visual conditioning frame; dropout 0.1 is applied on the CogVideoX conditioning and action denoising pathways to reduce overfitting.

\begin{table}[h]
\centering
\caption{\textbf{\modelname training hyperparameters} (CogVideoX-2B backbone).}
\label{tab:hparams}
\begin{tabular}{@{}ll@{}}
\toprule
\textbf{Optimizer} & AdamW \\
\quad Learning rate              & $1 \times 10^{-4}$ pretrain, $2 \times 10^{-5}$ finetune (constant) \\
\quad $\beta_1, \beta_2$          & 0.9, 0.95 \\
\quad Weight decay                & $1 \times 10^{-4}$ \\
\quad Gradient clipping (max norm) & 1 \\
\midrule
\textbf{Batch} & \\
\quad Per-GPU batch size          & 4 \\
\quad Number of GPUs              & 32 (4 $\times$ 8 H200) \\
\quad Effective batch size        & 128 \\
\quad Total gradient steps        & $75$k pretrain, $25$k finetune \\
\midrule
\textbf{Diffusion} & \\
\quad Parameterization            & $v$-prediction \\
\quad Schedule                    & CogVideoX trailing, zero terminal SNR \\
\quad Training timesteps $T$      & 1000 \\
\quad Joint-branch coupled $t$    & true \\
\quad Joint-vs-action-only mix prob $p_{\text{joint}}$ & 0.5 \\
\quad Action loss weight $\lambda_a$ & 1.0  \\
\quad Video loss weight $\lambda_v$ & 0.1  \\
\midrule
\textbf{Action chunk} & \\
\quad Chunk length $H$            & 16 \\
\quad Action dim $d_a$            & 20 \\
\quad Inference DDIM steps        & 1 \\
\midrule
\textbf{Regularization} & \\
\quad Dropout (cond/action paths) & 0.1 \\
\quad Image augmentations         & RandomResizeCrop, ColorJitter, GaussianBlur \\
\midrule
\textbf{Precision \& parallelism} & \\
\quad Mixed precision             & \texttt{bf16} \\
\quad Parallelism                 & FSDP, ZeRO-2 \\
\bottomrule
\end{tabular}
\end{table}

\paragraph{DreamZero baseline.} We reproduce DreamZero~\citep{ye2026dreamzero} with its default hyperparameters, full-finetuning a Wan2.2-TI2V-5B base model on 32 H200s with the same data as \modelname{} in each setting. On LIBERO we train to convergence ($100$k gradient steps). For the real-world tasks we follow DreamZero's default two-stage budget: $25$k pretraining steps on the shared video and action mixture followed by $25$k fine-tuning steps on the in-domain mixture ($50$k total). Unlike \modelname{}, DreamZero emits the full predicted video sequence at inference, the cost we benchmark in \S\ref{sec:efficiency}.

\paragraph{Training compute.} All models are trained on the same cluster. In the real-world setting \modelname{} uses $1.9$k H200-hours and DreamZero $2.4$k; on LIBERO \modelname{} uses $2.5$k, DreamZero $1.1$k, and Cosmos Policy $7.1$k.

\section{Dataset details}
\label{app:data}

Table~\ref{tab:data} lists the per-source statistics for our training mixture. The mixture has two tiers: cross-embodiment video sources used only for the video-diffusion objective, and in-domain target-platform data used for both the video and action objectives. ``Real hrs'' is the wall-clock duration of source footage; ``training hrs'' is the duration the model sees after the per-source effective FPS is applied. Total training-time duration is ${\sim}673$ hours of footage. The action-supervised slice (in-domain bimanual teleop, $6.02$ hrs of footage) drives half of every gradient step.

\begin{table}[h]
\centering
\caption{\textbf{Training data mixture.} Per-source wall-clock hours of footage (``real hrs''), source-native FPS, model-effective FPS, hours seen at training time after FPS adjustment, and per-batch sampling weight. Cross-embodiment video sources dominate raw hours but are throttled by sampling weight so that in-domain data drives each gradient step.}
\label{tab:data}
\resizebox{\columnwidth}{!}{
\begin{tabular}{@{}lcccccc@{}}
\toprule
Source & Type & Real hrs & Native FPS & Effective FPS & Training hrs & Batch weight \\
\midrule
\multicolumn{7}{@{}l}{\emph{Cross-embodiment video pretraining (video objective only)}} \\
\quad DROID~\citep{droid}                  & robot  & 131.8 & 15 & 15 & 131.8 & 0.027 \\
\quad Ego4D~\citep{ego4d}                  & human  & 32.1  & 30 & 30 & 32.1  & 0.020 \\
\quad Epic Kitchens~\citep{epickitchens}   & human  & 6.1   & 60 & 60 & 6.1   & 0.014 \\
\quad SSv2~\citep{ssv2}                    & human  & 108.5 & 12 & 12 & 108.5 & 0.020 \\
\quad Panda-70M~\citep{panda70m}           & web    & 165.4 & 30 & 30 & 165.4 & 0.127 \\
\quad BridgeData v2~\citep{bridgev2}        & robot  & 31.2  & 5  & 10 & 15.6  & 0.054 \\
\quad AgiBot-World~\citep{agibot}           & robot  & 412.6 & 30 & 60 & 206.3 & 0.182 \\
\midrule
\multicolumn{7}{@{}l}{\emph{In-domain target-platform data}} \\
\quad In-domain human task demonstrations  & human  & 0.75  & 30 & 30 & 0.75  & 0.055 \\
\quad In-domain bimanual teleop (three target tasks $+$ play) & robot$+$action & 6.02 & 30 & 30 & 6.02 & 0.500 \\
\midrule
\textbf{Total}  & & $895$ & & & $673$ & 1.000 \\
\bottomrule
\end{tabular}
}
\end{table}

\paragraph{Data mixture and routing.} The mixture follows Large Video Planner~\citep{lvp} and spans three categories (per-source hours and sampling weights in Table~\ref{tab:data}). Large-scale cross-embodiment robot data (DROID, BridgeData~v2, AgiBot-World) carries action labels in principle; we tested translating the BridgeData~v2 and AgiBot-World actions into our shared 20D representation (Appendix~\ref{app:action_align}) and routing them through the \emph{joint} and \emph{action} modes, but found that ignoring the labels and routing all three to the \emph{video} mode gave consistently better in-domain action performance, which is the configuration we use. Large-scale human and internet video (Ego4D, Epic Kitchens, SSv2, Panda-70M) carries no action labels and is also routed to the \emph{video} mode. The in-domain data is bimanual teleop at $30$\,Hz covering the three target tasks plus undirected ``play'' clips (an operator moving the robot through unscripted but reasonable motions, captioned after the fact); we additionally collect egocentric human videos of the target tasks and of out-of-distribution tasks, routed to the \emph{video} mode to bias the prior toward the deployment scene and embodiment. The teleop action space is 20D (per arm: 3D end-effector position, 6D rotation, 1D gripper), with actions being normalized following LBM~\citep{barreiros2025careful}.

\paragraph{Captions.}
Across all sources we use a per-clip language caption with maximum length 80 tokens after T5-XXL tokenization, which covers approximately 95.8\% of captions in the mixture. Longer captions are truncated to 80 tokens.

\paragraph{Filtering.}
For Panda-70M, Ego4D, and SSv2 we filter out clips with (i)~motion magnitude below a per-source threshold (static scenes), (ii)~detected text overlays exceeding 30\% of the frame area, and (iii)~aspect ratios outside the range $[0.5, 2.0]$. AgiBot-World and BridgeData v2 are used as released. In-domain teleop episodes flagged as failed by the operator are excluded.

\section{Action-space alignment}
\label{app:action_align}

All action-supervised sources are aligned to a shared 20D action space: per arm, 3D end-effector position $+$ 6D rotation~\citep{zhou2019continuity} $+$ 1D gripper aperture (left arm: 10D; right arm: 10D). BridgeData~v2 is single-arm; each trajectory is assigned to the left or right arm at random, and the unused arm is masked from the action loss. In-domain actions are chunk-relative: positions are offsets from the chunk's first action in the robot base frame, rotations are expressed relative to its orientation, and the gripper aperture is absolute.

\paragraph{Coordinate transforms.}
Cross-embodiment poses are expressed relative to the episode's first end-effector pose, in that pose's frame; AgiBot-World poses are then mapped into the target platform's axes by a fixed rotation. Rotations are then converted to the 6D continuous representation~\citep{zhou2019continuity}.

\paragraph{Gripper calibration.}
Per-source gripper signals are mapped to a normalized $[0, 1]$ aperture, with $0$ corresponding to fully closed and $1$ to fully open at the source-specific gripper extent. AgiBot-World's gripper signal is linearly rescaled between its measured open and closed values and clipped to $[0, 1]$; BridgeData~v2's gripper is binarized (open above $0.95$, closed below $0.05$, and intermediate values take the next binarized value).

\paragraph{Per-dimension normalization.}
Following Large Behavior Models~\citep{barreiros2025careful}, translational action dimensions are clipped to their per-dataset 2nd/98th percentiles and then linearly rescaled to $[-1, 1]$. This clipping is applied at training time only; at inference we rescale predictions back to the original action range before sending to the controller.

\paragraph{Temporal alignment.}
Cross-embodiment action streams are resampled to $30$\,Hz with one cubic spline fit jointly to all $20$ channels, including rotation and gripper. These streams were used only in the action-transfer experiments of Appendix~\ref{app:data}; the final model uses those sources as video only. Action chunks have length $H = 16$, corresponding to ${\sim}0.53$\,s of executed motion per inference call.

\section{Single-step inference: derivation, additional charts, and raw data}
\label{app:singlestep}

This appendix expands on \S\ref{sec:hf-investigation}: we derive the single-step DDIM update under our schedule and parameterization (\ref{app:derivation}); present the $v_\theta$-determinism measurements that quantify the regime collapse at the noise terminus (\ref{app:vt_variance}); report the full teacher-forced sweep and per-step iterative jitter (\ref{app:teacher_force}, \ref{app:raw_jitter}); rule out alternative explanations, including the choice of scheduler (\ref{app:alternatives}, \ref{app:scheduler_ablation}); compare to a Wan rectified-flow checkpoint trained on the same action data (\ref{app:wan_comparison}); document the spectral comparison between natural video and action chunks (\ref{app:fft}); and discuss the conditional-mean reduction as a deployment-time limitation of the chosen parameterization (\ref{app:cond_mean_limitation}).

\subsection{Derivation: 1-step DDIM under zero-terminal-SNR \texorpdfstring{$v$}{v}-prediction}
\label{app:derivation}

We follow the DDIM~\citep{song2020denoising} update with $v$-parameterization~\citep{salimans2022progressive}.
For a noise schedule $\{\bar{\alpha}_t\}_{t=0}^T$ and $v$-target
\begin{equation}
    \mathbf{v}_t \;=\; \sqrt{\bar{\alpha}_t}\,\boldsymbol{\epsilon} - \sqrt{1 - \bar{\alpha}_t}\,\mathbf{x}_0,
\end{equation}
the model $\mathbf{v}_\theta(\mathbf{x}_t, t, \mathbf{c})$ predicts $\mathbf{v}_t$ from the noisy sample $\mathbf{x}_t = \sqrt{\bar{\alpha}_t}\,\mathbf{x}_0 + \sqrt{1 - \bar{\alpha}_t}\,\boldsymbol{\epsilon}$ and conditioning $\mathbf{c}$.
The implied clean-sample estimate is
\begin{equation}
    \hat{\mathbf{x}}_0(\mathbf{x}_t, t, \mathbf{c}) \;=\; \sqrt{\bar{\alpha}_t}\,\mathbf{x}_t \;-\; \sqrt{1 - \bar{\alpha}_t}\,\mathbf{v}_\theta(\mathbf{x}_t, t, \mathbf{c}).
    \label{eq:appendix_x0_hat}
\end{equation}

We use the CogVideoX trailing schedule with effectively zero terminal SNR: $\bar{\alpha}_T \to 0$, so $\sqrt{\bar{\alpha}_T} \approx 0$ and $\sqrt{1 - \bar{\alpha}_T} \approx 1$, and the noisy sample at $t = T$ collapses to pure noise, $\mathbf{x}_T = \boldsymbol{\epsilon}$.
Substituting into \eqref{eq:appendix_x0_hat} gives
\begin{equation}
    \hat{\mathbf{x}}_0 \;=\; \sqrt{\bar{\alpha}_T}\,\boldsymbol{\epsilon} \;-\; \sqrt{1 - \bar{\alpha}_T}\,\mathbf{v}_\theta(\boldsymbol{\epsilon}, T, \mathbf{c}) \;\xrightarrow[\bar{\alpha}_T \to 0]{}\; -\mathbf{v}_\theta(\boldsymbol{\epsilon}, T, \mathbf{c}).
\end{equation}

The 1-step DDIM update from $t = T$ to $t = 0$ therefore depends on the noise input $\boldsymbol{\epsilon}$ only through $\mathbf{v}_\theta$. The $v$-target itself reduces to $\mathbf{v}_T = \sqrt{\bar{\alpha}_T}\,\boldsymbol{\epsilon} - \sqrt{1 - \bar{\alpha}_T}\,\mathbf{x}_0 \to -\mathbf{x}_0$ at $\bar{\alpha}_T = 0$, so the regression target at $t = T$ is $\boldsymbol{\epsilon}$-independent by construction; we verify the corresponding $\boldsymbol{\epsilon}$-invariance of $\mathbf{v}_\theta$ empirically in \S\ref{app:vt_variance}. Substituting into the 1-step update gives
\begin{equation}
    \hat{\mathbf{x}}_0 \;\approx\; -\mathbf{v}_\theta(\boldsymbol{\epsilon}, T, \mathbf{c}) \;\approx\; \mathbb{E}_{\mathbf{x}_0 \sim p(\mathbf{x}_0 \mid \mathbf{c})}[\mathbf{x}_0],
\end{equation}
i.e.\ the conditional mean of the action-chunk distribution given the observation--language conditioning $\mathbf{c}$. We discuss the deployment-time consequences of this conditional-mean reduction in \S\ref{app:cond_mean_limitation}.

\subsection{Determinism of \texorpdfstring{$\mathbf{v}_\theta$}{v\_theta} at \texorpdfstring{$t = T$}{t=T}}
\label{app:vt_variance}

\begin{figure}[h]
    \centering
    \includegraphics[width=0.85\linewidth]{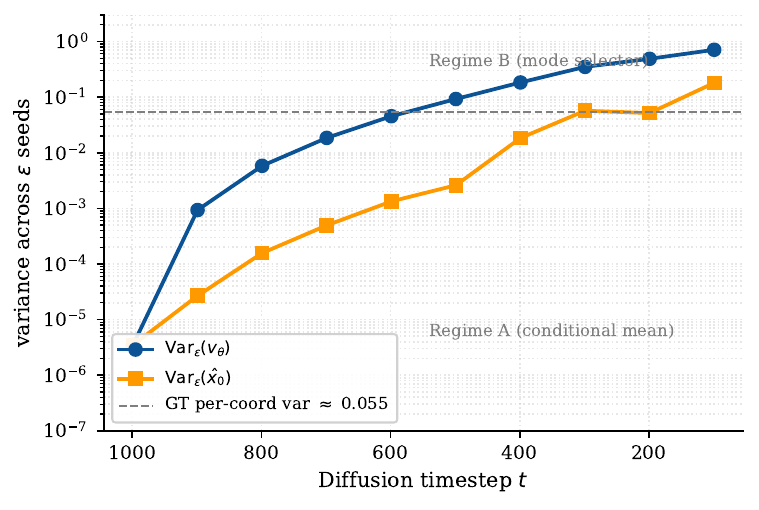}
    \caption{\textbf{Variance of $\mathbf{v}_\theta(\mathbf{x}_t, t, \mathbf{c})$ across $\boldsymbol{\epsilon}$ samples} ($N = 64$). Per-coordinate mean variance for both $\mathbf{v}_\theta$ and the implied $\hat{\mathbf{x}}_0$, plotted on a log scale across $10$ scheduled $t$ values. Reference bands show the spread of the empirical action distribution under matched conditioning. The variance at $t = T$ (regime A) is $\sim$5 orders below the spread of the conditional action distribution; for $t < T$ the model output is $\boldsymbol{\epsilon}$-coupled (regime B).}
    \label{fig:appendix_vt_variance}
\end{figure}

For each of $10$ diffusion timesteps spanning the schedule, we draw $N = 64$ i.i.d.\ $\boldsymbol{\epsilon}$ samples, construct $\mathbf{x}_t = \sqrt{\bar{\alpha}_t}\,\mathbf{x}_0 + \sqrt{1 - \bar{\alpha}_t}\,\boldsymbol{\epsilon}$ from a single fixed conditioning context $\mathbf{c}$, and compute the per-coordinate variance of $\mathbf{v}_\theta(\mathbf{x}_t, t, \mathbf{c})$ across the $64$ samples. The per-coordinate mean of this variance is plotted in Fig.~\ref{fig:appendix_vt_variance}. At $t = T$ the variance is $3.5{\times}10^{-6}$ per coordinate, five to six orders of magnitude below its value at any $t < T$ where the schedule retains $\boldsymbol{\epsilon}$-dependent target structure. The collapse is consistent with the analytic prediction in \S\ref{app:derivation}: at $\bar{\alpha}_T = 0$ the $v$-target reduces to $-\mathbf{x}_0$ and the regression problem decouples from $\boldsymbol{\epsilon}$.

\subsection{Teacher-forced vs.\ iterative inference}
\label{app:teacher_force}

\begin{figure}[h]
    \centering
    \includegraphics[width=0.65\linewidth]{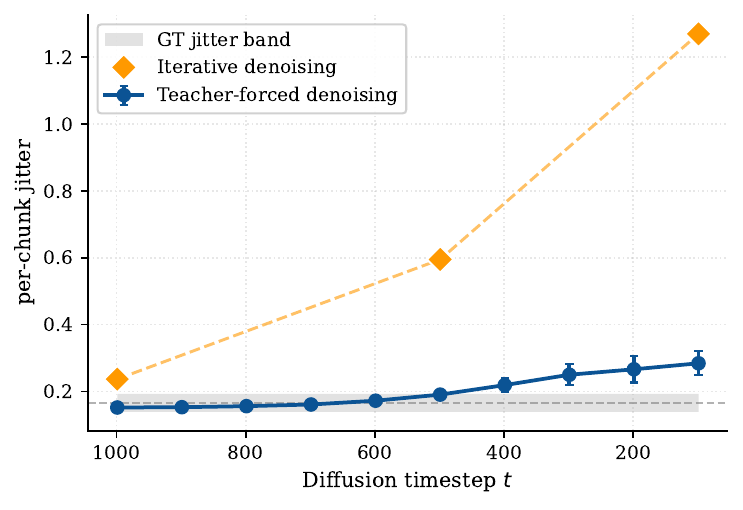}
    \caption{\textbf{Teacher-forced vs.\ iterative jitter at each $t$, $N = 50$ val chunks.} Teacher-forced (blue): single forward pass at each $t$ on in-distribution $\mathbf{x}_t$ from GT. Iterative diamonds (orange): $\hat{\mathbf{x}}_0^{(k)}$ from a $K = 10$ DDIM trajectory at $t \in \{999, 499, 99\}$. Gray band: GT chunk jitter $\pm 1\sigma$. Teacher-forced jitter rises monotonically but stays within $\sim\!2\times$ GT; iterative jitter grows to $\sim\!8\times$ GT, isolating exposure-bias compounding on top of the per-$t$ output bias.}
    \label{fig:appendix_teacher_force}
\end{figure}

We sweep $50$ held-out validation chunks across $10$ scheduled $t$ values. For each chunk and each $t$, the teacher-forced input is $\mathbf{x}_t = \sqrt{\bar{\alpha}_t}\,\mathbf{x}_0 + \sqrt{1-\bar{\alpha}_t}\,\boldsymbol{\epsilon}$, a single forward pass yields $\mathbf{v}_\theta$ from which $\hat{\mathbf{x}}_0$ is computed via Eq.~\ref{eq:appendix_x0_hat}; we report mean and standard deviation across chunks of the per-chunk RMS second-difference of $\hat{\mathbf{x}}_0$ at each $t$. Teacher-forced jitter rises monotonically with decreasing $t$, from $0.92\times$ the GT band at $t = T$ to $1.71\times$ the GT band at $t = 99$, despite by-construction in-distribution inputs at each $t$ (Fig.~\ref{fig:appendix_teacher_force}, blue). This rules out off-manifold inputs as the cause of high-frequency growth: the model's outputs themselves carry high-frequency mass at low $t$, consistent with the mis-tuned video frequency prior of \S\ref{sec:hf-investigation}. The iterative trajectory amplifies the same low-$t$ residuals into per-step compounding (Table~\ref{tab:appendix_hf_full}, columns 3--4).

\subsection{Alternative explanations ruled out}
\label{app:alternatives}

We tested four alternatives to the video-pretraining account of \S\ref{sec:hf-investigation} by direct measurement. \emph{AdaLN timestep coupling} is not the cause: an action-only ablation that removes the shared AdaLN modulation between video and action tokens exhibits the same monotonic jitter growth as $t \to 0$. \emph{Decoupled-timestep sampling} does not help either: variants that draw the action timestep separately from the video timestep, following Cosmos Policy~\citep{kim2026cosmospolicyfinetuningvideo} (hybrid log-normal/uniform), DreamZero~\citep{ye2026dreamzero} (action-side Beta), and UWM~\citep{zhu2025unified} (independent timesteps), all retain the growth on iterative inference. The remaining two candidates, \emph{scheduler choice} and \emph{$v$-prediction parameterization}, are examined in \S\ref{app:scheduler_ablation} and \S\ref{app:wan_comparison}; neither removes it. Each of the four would apply equally to from-scratch action diffusion policies, which show no such pathology, leaving the video pretraining as the distinguishing variable.

\subsection{Scheduler ablation}
\label{app:scheduler_ablation}

We tested DDIM (default), UniPC, and a zero-shift variant of the schedule across $K \in \{1, 10, 30\}$ inference steps. High-frequency jitter was present in every multi-step configuration and absent at $K = 1$ in every configuration; the qualitative pattern matches the iterative-trajectory column of Table~\ref{tab:appendix_hf_full}. The choice of multi-step sampler does not eliminate high-frequency growth. On the robot, jitter is visible from $K = 2$ and grows with $K$ unless the chunk is smoothed post hoc; on LIBERO, $1$-step and $10$-step inference reach comparable success after smoothing. Filtered $10$--$30$-step inference was usable on early checkpoints but not on the final model, so the high-frequency bias did not shrink with training.

\subsection{Raw per-step jitter measurements}
\label{app:raw_jitter}

Table~\ref{tab:appendix_hf_full} reports the full $10$-point teacher-forced sweep behind Fig.~\ref{fig:appendix_teacher_force}, with mean and standard deviation across $N = 50$ val chunks at each scheduled $t$. The iterative-trajectory column is taken from a single representative DDIM rollout (matching the markers in Fig.~\ref{fig:appendix_teacher_force}); per-context variance of the iterative numbers is qualitatively similar.

\begin{table}[h]
\centering
\caption{\textbf{Teacher-forced sweep and iterative trajectory jitter at each scheduled $t$.} Jitter is the per-chunk RMS of the second finite difference of $\hat{\mathbf{x}}_0$ across the $16$-step horizon. GT band: $0.166 \pm 0.027$.}
\label{tab:appendix_hf_full}
\begin{tabular}{@{}cccc@{}}
\toprule
DDIM step $k$ & $t$ & TF $\|\hat{x}_0\,\mathrm{jit}\|$ (mean $\pm$ std, 50 chunks) & Iterative $\|\hat{x}_0^{(k)}\,\mathrm{jit}\|$ (1 trajectory) \\
\midrule
0 & 999 & $0.152 \pm 0.011$ & 0.237 \\
1 & 899 & $0.153 \pm 0.011$ & --    \\
2 & 799 & $0.156 \pm 0.012$ & --    \\
3 & 699 & $0.161 \pm 0.012$ & --    \\
4 & 599 & $0.172 \pm 0.013$ & --    \\
5 & 499 & $0.190 \pm 0.014$ & 0.595 \\
6 & 399 & $0.219 \pm 0.021$ & --    \\
7 & 299 & $0.250 \pm 0.031$ & --    \\
8 & 199 & $0.266 \pm 0.039$ & --    \\
9 &  99 & $0.284 \pm 0.036$ & 1.27  \\
\bottomrule
\end{tabular}
\end{table}

\subsection{Wan rectified-flow comparison}
\label{app:wan_comparison}

To test whether the high-frequency pathology is specific to the CogVideoX backbone or a property of the broader class of repurposed video-generation models, we trained a Wan-2.1 video backbone on the same action data with identical action tokenization, observation conditioning, and language conditioning, but the rectified-flow $v$-prediction objective and \texttt{FlowMatchScheduler} that Wan uses for its native video pretraining. Concretely, the noisy sample is $\mathbf{x}_\sigma = (1 - \sigma)\,\mathbf{x}_0 + \sigma\,\boldsymbol{\epsilon}$, the regression target is $\mathbf{v}_\sigma = \boldsymbol{\epsilon} - \mathbf{x}_0$, and inference uses Euler updates $\mathbf{x}_{\sigma'} = \mathbf{x}_\sigma + \mathbf{v}_\theta\,(\sigma' - \sigma)$, with $\sigma_{\max} = 1$ and $\sigma_{\min} \approx 0.003$. The crucial distinction from CogVideoX's schedule is that the target at the noise terminus is $\mathbf{v}_{\sigma_{\max}} = \boldsymbol{\epsilon} - \mathbf{x}_0$, which retains explicit $\boldsymbol{\epsilon}$-dependence and so does \emph{not} structurally collapse to an $\boldsymbol{\epsilon}$-independent regression problem.

\begin{figure}[h]
    \centering
    \includegraphics[width=\linewidth]{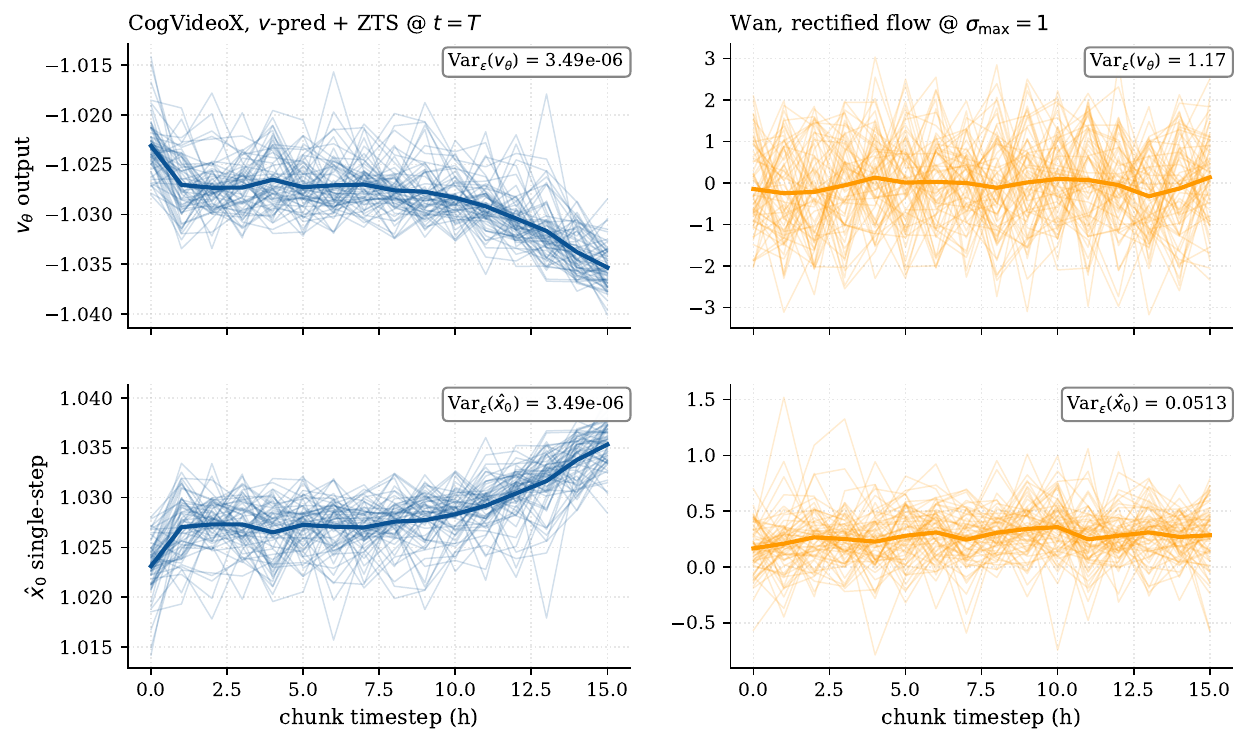}
    \caption{\textbf{Parameterization specificity at the noise terminus.} 64-seed traces of $\mathbf{v}_\theta$ (top) and the implied $\hat{\mathbf{x}}_0$ (bottom) for the highest-variance action dim, with all other inputs held fixed. Left (blue): \modelname (CogVideoX, $v$-pred + ZTS) at $t = T$ ($\bar{\alpha}_T = 0$): $\mathrm{Var}_{\boldsymbol{\epsilon}}\,\mathbf{v}_\theta = 3.5{\times}10^{-6}$, traces collapse onto a single curve. Right (orange): Wan rectified flow at $\sigma_{\max} = 1$: $\mathrm{Var}_{\boldsymbol{\epsilon}}\,\mathbf{v}_\theta = 1.17$, traces span the action range. Five-and-a-half orders of magnitude separate the two parameterizations on this same metric.}
    \label{fig:appendix_param_compare}
\end{figure}

\begin{figure}[h]
    \centering
    \includegraphics[width=0.95\linewidth]{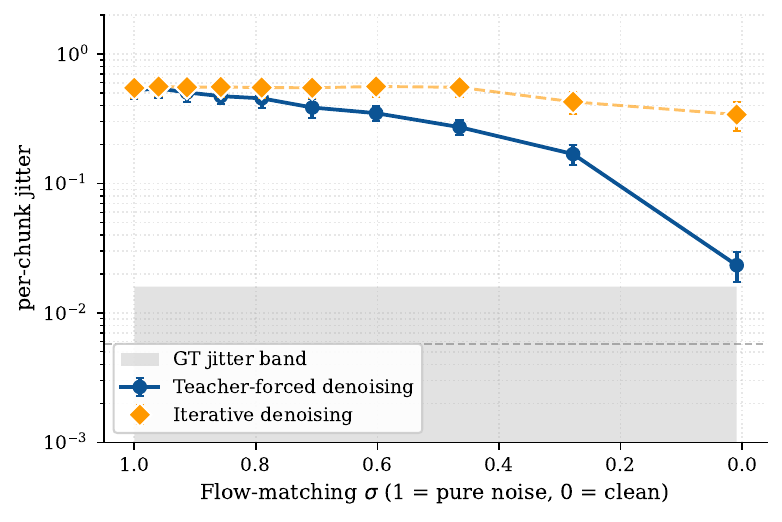}
    \caption{\textbf{Wan rectified-flow analog of Fig.~\ref{fig:appendix_teacher_force}.} Teacher-forced single-step jitter across $\sigma$ (blue) collapses with $\sigma$ as $\hat{\mathbf{x}}_0 = \mathbf{x}_\sigma - \sigma\,\mathbf{v}_\theta$. The iterative trajectory (orange diamonds) stays elevated throughout and converges to $\sim$59$\times$ the GT band at $\sigma_{\min}$. Unlike CogVideoX, Wan has no $\sigma$ at which the regression target structurally decouples from $\boldsymbol{\epsilon}$, so the corresponding gate around the mis-tuned low-$\sigma$ frequency prior is unavailable.}
    \label{fig:appendix_wan_tf_iter}
\end{figure}

We measure the parameterization signature directly: across $N = 64$ $\boldsymbol{\epsilon}$ samples at $\sigma = \sigma_{\max} = 1$, the per-coordinate variance of $\mathbf{v}_\theta$ is $1.17$, an $\mathcal{O}(1)$ value consistent with the $\boldsymbol{\epsilon}$-dependent target (cf.\ \modelname's $3.5{\times}10^{-6}$ at $t = T$ in \S\ref{app:vt_variance}). Fig.~\ref{fig:appendix_param_compare} contrasts the two backbones at the noise terminus across $64$ seed traces over the chunk horizon. Fig.~\ref{fig:appendix_wan_tf_iter} reports a teacher-forced and iterative sweep on $50$ held-out validation chunks across $10$ scheduled $\sigma$ values: teacher-forced jitter falls from $0.524$ at $\sigma_{\max} = 1$ to $0.023$ at $\sigma_{\min}$, scaling as expected from $\hat{\mathbf{x}}_0 = \mathbf{x}_\sigma - \sigma\,\mathbf{v}_\theta$ (the prediction residual is multiplied by $\sigma$ which goes to zero), but the iterative trajectory does \emph{not} collapse with $\sigma$: it stays near $0.55$ across the schedule and the converged single-step output at $\sigma_{\min}$ has jitter $0.34$, $\sim$59$\times$ the Wan GT band. The Wan single-step output at $\sigma_{\max}$ is also far from the GT band ($\sim$91$\times$). Neither single-step nor iterative inference under rectified flow recovers the structural escape hatch that \modelname enjoys at $\bar{\alpha}_T = 0$. We kept Wan's native objective deliberately to isolate the mechanism; initial experiments replacing it were unstable, and whether doing so recovers single-step inference remains open.

\subsection{Conditional-mean reduction}
\label{app:cond_mean_limitation}

The structural property of \S\ref{para:hf_theory} that makes single-step inference well-defined ($\bar{\alpha}_T = 0$ collapsing the $v$-target to $-\mathbf{x}_0$) also forces the trained model to act as a conditional-mean estimator at the noise terminus. Eq.~\ref{eq:single_step} reads $\hat{\mathbf{x}}_0 = -\mathbf{v}_\theta(\boldsymbol{\epsilon}, T, \mathbf{c})$, and the determinism measurement Var$_{\boldsymbol{\epsilon}}\,\mathbf{v}_\theta = 3.5{\times}10^{-6}$ at $t = T$ confirms that this output is essentially independent of $\boldsymbol{\epsilon}$. Single-step inference therefore returns $\mathbb{E}[\mathbf{x}^a_0 \mid \mathbf{c}]$ rather than samples from $p(\mathbf{x}^a_0 \mid \mathbf{c})$. Deployment queries only $t = T$, where the objective reduces to regression onto $\mathbf{x}^a_0$. We did not train at $t = T$ alone, but timestep samplers biased toward high noise, including those of DreamZero, Cosmos Policy, and UWM, had no significant effect on performance.

Recovering distributional sampling within this framework requires changing one of three components. (i)~Parameterization: retaining $\boldsymbol{\epsilon}$-dependence at the noise terminus (rectified flow as in App.~\ref{app:wan_comparison}, or non-zero terminal-SNR $\boldsymbol{\epsilon}$-prediction as in Diffusion Policy~\citep{chi2023diffusion}) and accepting iterative inference with its high-frequency cost. (ii)~Conditioning: an auxiliary discrete mode-head that breaks the conditional-mean tie before the single-step head. (iii)~Inference recipe: sampling multiple chunks from a learned latent prior over a coarse action class and decoding each via the single-step head. The relative tradeoffs are an empirical question we leave to future work. Our preferred first remedy, ahead of these three, is more detailed text conditioning, which can act as a mode selector.

\subsection{Symmetric-target probe}
\label{app:symmetric_probe}

To test whether the conditional mean averages between modes in practice, we placed two identical croissants in front of the robot, varied their spacing and placement across the workspace, and gave no disambiguating language. \modelname{} succeeded in $8/10$ rollouts, matching an $8/10$ single-croissant control. Every reach committed to one croissant ($7$ right, $3$ left); no rollout reached between the objects or failed to commit to a grasp. Across the $30$ in-distribution and $15$ out-of-distribution real-world rollouts and the four LIBERO suites we likewise observed no averaging, so none of the remedies above was needed.

\subsection{Spectral comparison: natural video vs.\ action chunks}
\label{app:fft}

To support the mechanism claim of \S\ref{sec:hf-investigation} we measure the spectral content of the two distributions the backbone is trained on: the natural-video pixels of its pretraining corpus and the action chunks of its \modelname finetuning. Both signals are sampled from the same teleoperation episodes (the video pixels are the same recording the actions are aligned to, at the native $30$\,Hz capture rate), so video and action share the same Nyquist frequency and frequency grid. We compute the FFT of unit-energy-normalized 16-step chunks (the model's prediction horizon) and 256-step Hann windows (long-form spectrum) for both modalities and average the resulting power spectra; results are summarized in Table~\ref{tab:fft_bands}.

\begin{table}[h]
\centering
\caption{\textbf{Spectral energy distribution: natural video (pixels) vs.\ action chunks.} Unit-energy-normalized power, averaged per modality. ``Frac.\ above $f$'' is the fraction of total energy above the listed frequency. Long-form bands from a $256$-step Hann window FFT; chunk-horizon $f_{50}$/$f_{95}$ from a $16$-step rectangular FFT.}
\label{tab:fft_bands}
\begin{tabular}{@{}lcc@{}}
\toprule
Quantity & Action chunks & Natural video pixels \\
\midrule
$f_{50}$ (chunk horizon, 16 steps) & 0.00\,Hz & 1.875\,Hz \\
$f_{95}$ (chunk horizon, 16 steps) & 1.875\,Hz & 7.50\,Hz \\
Frac.\ above 3.75\,Hz (chunk horizon) & 0.014 & 0.122 \\
Frac.\ above 7.5\,Hz (chunk horizon) & 0.008 & 0.048 \\
$f_{95}$ (long-form, 256 steps) & 0.352\,Hz & 1.992\,Hz \\
Frac.\ in $[0.75, 1.5]$\,Hz (long-form) & 0.011 & 0.080 \\
Frac.\ in $[1.5, 3.0]$\,Hz (long-form) & 0.004 & 0.046 \\
Frac.\ in $[3.0, 7.5]$\,Hz (long-form) & 0.001 & 0.023 \\
\bottomrule
\end{tabular}
\end{table}

Across both horizons, the $f_{95}$ of natural video is $4$--$6\times$ that of action chunks, and roughly a third of video energy lies above the action $f_{95}$ at the chunk horizon. In the noise schedule's low-$t$ regime, where the model denoises small additive noise atop a near-clean signal, this difference rewards different behavior on the two distributions: on natural video, the model is rewarded for synthesizing high-frequency detail (textures, edges, fine motion); on action chunks, the GT signal does not contain that energy and the same low-$t$ behavior produces out-of-distribution high-frequency content in $\hat{\mathbf{x}}_0$.

\textbf{Caveats.} (i)~Action chunks are normalized by per-timestep percentile scaling at position dims; 6D rotation and gripper dims remain unnormalized. Long-form action spectra cannot apply per-timestep scaling at horizons $\neq 16$ and so we use chunk-relative subtraction at position dims with no percentile scaling; this affects absolute spectral magnitude but not shape, and is moot under the unit-total-energy normalization used here. (ii)~Video is reported in pixel space ($128{\times}128$ RGB at native fps), not in CogVideoX's VAE latent space: the temporal-stride-4 VAE acts as a low-pass filter, so latent-space high-frequency energy is upper-bounded by pixel-space high-frequency energy, and the high-frequency claim in pixels is a conservative claim in latent space.